\documentclass[pdflatex,sn-basic, referee]{sn-jnl}% Math and Physical Sciences Numbered Reference Style
\usepackage{graphicx}%
\usepackage{multirow}%
\usepackage{amsmath,amssymb,amsfonts}%
\usepackage{amsthm}%
\usepackage{mathrsfs}%
\usepackage[title]{appendix}%
\usepackage{xcolor}%
\usepackage{textcomp}%
\usepackage{manyfoot}%
\usepackage{booktabs}%
\usepackage[utf8]{inputenc}
\usepackage[english]{babel}
\usepackage{enumerate}
\usepackage{mathtools}
\usepackage{comment}
\usepackage{bbm}
\usepackage{booktabs}
\usepackage{pifont}
\usepackage{algorithm2e}
\usepackage{subcaption}
\usepackage[normalem]{ulem}
\usepackage{hyperref}
\usepackage{lscape}
\usepackage{csquotes}
\usepackage{multirow}
\usepackage{rotating}
\usepackage{makecell}
\usepackage{tabularx}
\usepackage{dcolumn}
\usepackage{units}
\usepackage{array}
\usepackage{float}

\newcommand{\R}{\mathbb{R}}

\newcommand{\pPP}{\boldsymbol{\mathcal{P}}}
\newcommand{\zZ}{\mathbf{Z}}
\newcommand{\aA}{\mathbf{A}}
\newcommand{\aAA}{\boldsymbol{\mathcal{A}}}
\newcommand{\dD}{\mathbf{D}}
\newcommand{\wW}{\mathbf{W}}
\newcommand{\lL}{\mathbf{L}}
\newcommand{\iI}{\mathbf{I}}
\newcommand{\pP}{\mathbf{P}}
\newcommand{\cC}{\mathbf{C}}
\newcommand{\xX}{\mathbf{X}}
\newcommand{\sS}{\mathbf{S}}
\newcommand{\bB}{\mathbf{B}}

\newcommand{\kK}{\mathbf{K}}
\newcommand{\zz}{\mathbf{z}}
\newcommand{\cc}{\mathbf{c}}

\newcommand{\mm}{\mathbf{m}}
\newcommand{\rr}{\mathbf{r}}
\newcommand{\xx}{\mathbf{x}}
\newcommand{\ttt}{\mathbf{t}}
\newcommand{\EE}{\mathbb{E}}

\newcommand{\N}{\mathbb{N}}
\theoremstyle{thmstyleone}%
\theoremstyle{thmstyletwo}%
\newcommand{\eg}{\textit{e}.\textit{g}., }
\newcommand{\ie}{\textit{i}.\textit{e}., }

\theoremstyle{thmstylethree}%
\newtheorem{definition}{Definition}%

\begin{document}

\title[Embeddings on Multiplex Networks]{Embedding Learning on Multiplex Networks for Link Prediction}


\author*[1]{\fnm{Orell} \sur{Trautmann}}\email{orell.trautmann@uni-rostock.de}

\author[1,2,3]{\fnm{Olaf} \sur{Wolkenhauer}}\email{olaf.wolkenhauer@uni-rostock.de}

\author[4]{\fnm{Cl\'{e}mence} \sur{R\'eda}}\email{reda@bio.ens.psl.eu}

\affil[1]{\orgdiv{Institute of Computer Science}, \orgname{University of Rostock}, \orgaddress{%\street{Ulmenstrasse 69, Haus 3}, 
\city{Rostock}, \postcode{18057}, \country{Germany}}}

\affil[2]{\orgname{Leibniz-Institute for Food Systems Biology at Technical University Munich}, \orgaddress{\city{Freising}, \postcode{85354}, \country{Germany}}}

\affil[3]{\orgdiv{Stellenbosch Institute of Advanced Study}, \orgname{Wallenberg Research Centre}, \orgaddress{\city{Stellenbosch}, \postcode{7602}, \country{South Africa}}}

\affil[4]{\orgdiv{BioComp}, \orgname{Institut de biologie de l’Ecole normale supérieure (IBENS), Ecole normale supérieure, CNRS, INSERM, PSL Université}, \orgaddress{\city{Paris}, \postcode{75005}, \country{France}}}


\abstract{Over the past years, embedding learning on networks has shown tremendous results in link prediction tasks for complex systems, with a wide range of real-life applications. Learning a representation for each node in a knowledge graph allows us to capture topological and semantic information, which can be processed in downstream analyses later. In the link prediction task, high-dimensional network information is encoded into low-dimensional vectors, which are then fed to a predictor to infer new connections between nodes in the network. As the network complexity (that is, the numbers of connections and types of interactions) grows, embedding learning turns out increasingly challenging. This review covers published models on embedding learning on multiplex networks for link prediction. First, we propose refined taxonomies to classify and compare models, depending on the type of embeddings and embedding techniques. Second, we review and address the problem of reproducible and fair evaluation of embedding learning on multiplex networks for the link prediction task. Finally, we tackle evaluation on directed multiplex networks by proposing a novel and fair testing procedure. This review constitutes a crucial step towards the development of more performant and tractable embedding learning approaches for multiplex networks and their fair evaluation for the link prediction task. We also suggest guidelines on the evaluation of models, and provide an informed perspective on the challenges and tools currently available to address downstream analyses applied to multiplex networks.}


\maketitle

\section{Introduction}

Networks have become a standard tool for modelling complex systems, due to their ability to abstract interactions between entities~\citep{ComplexSystems}, and compile information into knowledge graphs. Networks (or graphs) are a diagrammatic representation of complex systems, which depict entities as nodes (or vertices) and their interrelations as edges~\citep{StructureOfCompNet}. While classical network approaches have achieved tremendous success in modelling real-world problems~\citep{MultilayerReview2014}, they do not adequately represent the variety of information. As such, over the past two decades, multiplex networks have grown in popularity~\citep{MultilayerReview2014, Boccaletti2014TheSA, BianconiMulNetBook}. Multiplex networks are a collection of networks that share the same nodes but have different edge combinations, thereby representing a generalization of graphs. Multiplex networks are especially useful for representing interactions or connections between different entities from various perspectives, which is why they are sometimes also referred to as multi-view networks~\citep{MVE}.

Multiplex networks have numerous real-life applications; for example, they can represent user interactions across social media platforms~\citep{SocialMultiplex,OnlineSocialNetworksMultiplex} or protein-protein interactions across tissues~\citep{OhmNet}. More specifically, in the example on social media, multiplex networks can be used to model cross-platform social relationships. Such a network would consist of multiple layers, each corresponding to an online social network~\citep{SocialMultiplex} on a singular platform (\eg Facebook, Instagram, etc.). The social networks themselves are constructed in a way that the nodes correspond to people (their accounts) and the edges between them correspond to a friendship (or the action of following a user
). Since people may have accounts across different social media platforms, these networks are interconnected, forming a network of networks~\citep{Boccaletti2014TheSA, MultilayerReview2014, BianconiMulNetBook}. Similarly, in the example of tissue-specific protein-protein interactomes, the nodes correspond to proteins, and the edges represent their pairwise interactions, which may differ across the tissues considered, \ie, the layers. This multiplex network, therefore, characterizes the proteins' tissue-dependent behaviors.

These networks are not only useful as a way to summarize data, they also play a crucial role in many machine learning and data mining tasks~\citep{Gheche-Frossard}. Networks are intrinsically combinatorial objects without a vector space~\citep{zooGuide}. Network embeddings map these objects to a convenient latent space, which allows us to transform the information contained within into a machine learning-suitable representation. This process is also known as representation learning~\citep{zooGuide}, by essentially vectorizing nodes~\citep{MUNEM}. As the use of multiplex networks in describing real-world systems has increased, efforts to generalize graph embeddings in the multiplex context have also grown. The difficulty here lies in producing vector representations for each node utilizing the additional information from other network layers. The vectorized forms of the networks are then used in downstream tasks such as node classification~\citep{MEGAN}, community detection~\citep{Multi-node2vec}, network reconstruction~\citep{RMNE}, and link prediction~\citep{MNI}.

In link prediction, the goal is to infer the presence or the absence of a connection between two nodes~\citep{linkpredictionfaireval, LinkPredWeightedEval}. For the cross-platform example, this could be understood as suggesting new friends on Facebook, given the friendship information from all social media platforms. Similarly, link prediction allows us to uncover novel protein-protein interactions in protein-protein interactomes. The challenge is that the prediction relies on the view, or the layer of connections, so the embeddings must also capture both the differences and similarities across views or layers of the multiplex network.

Many models have been proposed for embedding learning on networks for link prediction~\citep{PMNE, OhmNet, MNE, RMNE, MultiplexSAGE}. Yet, to our knowledge, apart from a very brief review on multiplex network embeddings for clustering~\citep{MultiplexClusterReview}, there has been no substantial review of multiplex network embedding methods. Furthermore, the various types of latent representation vectors used in the literature have not yet been classified.

Our contributions to the field are as follows: first and foremost, in Section \ref{RepTaxonomy} we introduce a first representation or embedding taxonomy for multiplex network embeddings. Second, in Section \ref{taxonomy}, we extend and refine an existing classification of embedding techniques on single-layer network embeddings~\citep{zooGuide} to multiplex networks, incorporating an additional aggregation category~\citep{PMNE}. Then, this method taxonomy is leveraged to provide a detailed, yet general, description of the pros and cons of each type of embedding technique. In particular, we will focus on approaches that permit the integration of information from different views of the multiplex networks. Third, we discuss and propose guidelines for the evaluation problem on link prediction in Section \ref{metrics}. Moreover, we aim to address the evaluation on directed multiplex networks by introducing a new testing procedure.

\section{Definitions and Preliminaries on Networks}\label{definitions}

Here, we introduce the relevant terminology and notation for the next sections. For a detailed description of multiplex networks, the reader is referred to~\cite{Boccaletti2014TheSA,MultilayerReview2014,BianconiMulNetBook}. We begin with a brief overview of the relevant terminology and notation for ``simple" graphs, which we will use recurrently throughout this review. 

A graph $G$ is described by an ordered pair $(V, E)$, where $V:=\{v_1, v_2, ..., v_N\}$ is the set of nodes (or vertexes), $E\subseteq V\times V$ is the set of edges between pairs of nodes. We distinguish between directed and undirected graphs. For undirected graphs, the pair $(v_1,v_2)\in E$ is unordered. In case of directed graphs, these pairs are indeed ordered, which means for $(v_1,v_2)\in E$, $(v_2,v_1)$ need not necessarily be an element of $E$. The edges may additionally be weighted by the function $w:V\times V\rightarrow\R_+$. An unweighted graph can be understood as a weighted network with values in $\{0,1\}$. Finally, the information from the graph can be represented by the adjacency matrix $\aA=\left(w(v,u)\right)_{(v,u)\in V\times V}$. The degree of a node $v$ is the sum of the weights of the edges connected to it, \ie $\deg(v)=\sum_{u\in V}w(u,v)$. The degree matrix of the network is the diagonal matrix $\dD=\text{diag}(\deg(v_1),..., \deg(v_N))$. These matrices play an important role in neural network-based methods that we will cover in Section~\ref{neraul-network}.

\begin{figure}[ht]
    \begin{subfigure}[b]{0.32\textwidth}
         \centering
         \includegraphics[width=.9\textwidth]{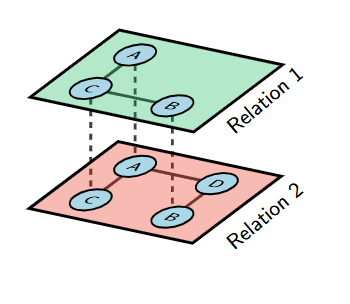}
         \caption{Multiplex network with two layers\\}
         \label{fig:multiplex2layers}
     \end{subfigure}
     \hfill
     \begin{subfigure}[b]{0.32\textwidth}
         \centering
         \includegraphics[width=.8\textwidth]{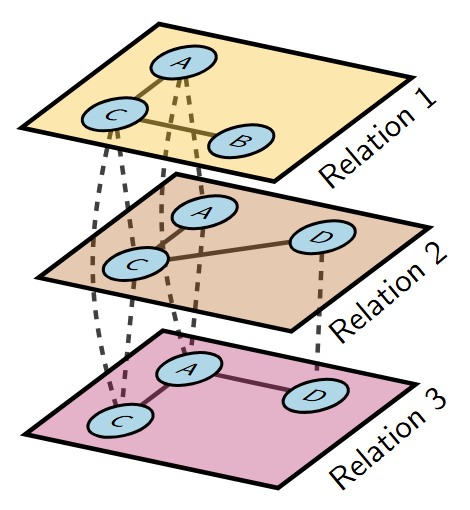}
         \caption{Multiplex network with three layers\\}
         \label{fig:multiplex3layers}
     \end{subfigure}
     \hfill
     \begin{subfigure}[b]{0.32\textwidth}
         \centering
         \includegraphics[width=.8\textwidth]{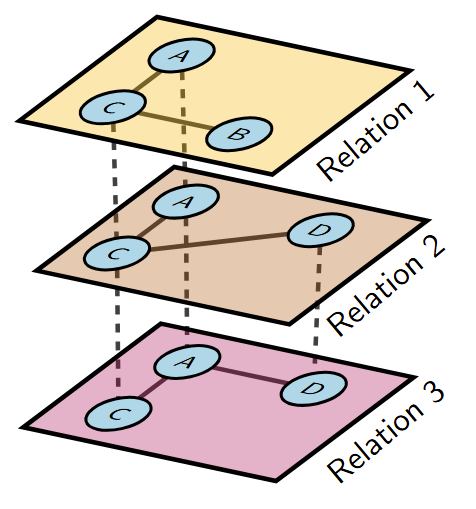}
         \caption{The same multiplex network  
         with three layers in another style}
         \label{fig:multiplex3layersOther}
     \end{subfigure}
     \hfill
    \caption{Examples of multiplex networks.     Each layer corresponds to a specific relation, or view, in the network. The dashed lines represent interlayer connections between replica nodes (same entity represented on different layers), which are the only allowed type of interlayer edges in multiplex networks, whereas solid lines match
    intralayer connections. Figures \ref{fig:multiplex3layers} and \ref{fig:multiplex3layersOther} both show
    the same network using two different styles: with an edge to every replica node or simply edges to connect the layers from the top layer to the bottom layer.}
    \label{fig:twographs}
\end{figure}

In prior works, many authors use the terms ``single-layer network'' or ``monoplex network'' to make a clear distinction between networks featuring a single view, and those with several views or layers~\citep{mathematicalFormulationMN, Boccaletti2014TheSA, MultilayerReview2014} (Figure~\ref{fig:twographs}). Two mathematical notations for multiplex networks seem to prevail in the literature. The notation from~\cite{BianconiMulNetBook} and~\cite{Boccaletti2014TheSA} emerged from the realm of physics, whereas~\cite{MultilayerReview2014} introduces notation more closely related to the literature in computer science. As it is more common in the methods we have reviewed, we will rely on the notation in~\cite{BianconiMulNetBook} and~\cite{Boccaletti2014TheSA}.

\begin{definition}[\textit{Multiplex Network}]\label{MultiplexDef}
    A multiplex network is a pair
    $\mathcal{M}=(Y,\mathcal{G})$, where $Y=\{1,...,M\}$ is the set of layers, and $\mathcal{G}=\{G_\alpha\}_{\alpha=1}^{M}$ is a family of graphs $G_\alpha = (V_\alpha,E_\alpha)$ with $\bigcap_{\alpha\in Y}V_\alpha\neq\emptyset$. Graph $G_\alpha$ on layer $\alpha$ can be undirected, directed, and/or weighted.
\end{definition}

Nodes representing the same entity in different layers are called replica nodes~\citep{BianconiMulNetBook}. To distinguish between replica nodes, we will use the notation $v_i^{[\alpha]}$ for entity $v_i\in V$ on layer $\alpha\in Y$. Edges within a layer, \ie the elements of $(E_{\alpha})_{\alpha}$, are called intralayer edges. On the other hand, only connections between replica nodes are allowed when considering edges crossing layers. Edges between replica nodes of different layers are referred to as interlayer edges. We say the multiplex is node-aligned if $V_\alpha = V$ for all $\alpha\in Y$~\citep{MultilayerReview2014}. As depicted in Figure \ref{fig:twographs}, it may be the case that nodes in some of the layers are missing. Without a loss of generality, one can assume that all layers feature the same set of nodes $V$, except that some of the nodes will be isolated depending on the layer~\citep{MultilayerReview2014}.

Similarly to graphs, multiplex networks can be represented by their supra-adjacency matrix $\aAA$.~\footnote{Usually, the supra-adjacency matrix is introduced as the flattened adjacency tensor of $\mathcal{M}$~\citep{MultilayerReview2014, Boccaletti2014TheSA, BianconiMulNetBook}. This explanation is omitted as it is not used in our review.} For the family of adjacency matrices $\{\aA_\alpha\}_{\alpha=1}^{M}$ corresponding to $\mathcal{G}$, and if $N=\#V$ be the number of distinct nodes, then the corresponding supra-adjacency matrix is given by
\begin{equation}\label{generalSupraAdjMat}
    \aAA=\begin{pmatrix}
        \aA_1 & \iI_{N} & \hdots & \iI_{N}\\
        \iI_{N} & \aA_2 & \ddots &\vdots\\
        \vdots & \ddots & \ddots & \iI_{N}\\
        \iI_{N} & \hdots & \iI_{N} & \aA_M
    \end{pmatrix}\;,
\end{equation}

\noindent where $\iI_{N}$ is the identity matrix of dimension $N \times N$, with zeroes everywhere but on the diagonal, where all coefficients are set to $1$. This structure will reappear in Section \ref{random-walk}, which discusses random walk-based methods. With this notation in mind, we can now introduce embedding representation and method taxonomies.

\section{Representation and Method Taxonomies}\label{Taxonomy}

Taxonomies are essential for classifying models, leading to improved understanding and comparisons between methods. To our knowledge, the only other review on embedding techniques for multiplex networks is~\cite{MultiplexClusterReview}, and focuses on downstream clustering tasks. The authors differentiate between embedding approaches on single-layer networks, which are applied to each layer individually and then merge embeddings across layers to obtain a single embedding vector per node, and embedding learning on multiplex networks. However, they did not dwell further on the differences between methods.~\cite{WhereFuseInformation} gave a more detailed report on graph neural networks for  representation learning on multiplex networks by introducing a taxonomy on information fusion for graph neural networks (GNN). Nevertheless, this taxonomy cannot be straightforwardly extended to multiplex network embeddings. Therefore, we provide a general taxonomy for multiplex network embedding methods.

Additionally, we establish that the latent representation embedding vectors obtained in multiplex networks are more nuanced than in monoplex networks. The different types of representation vectors are sometimes mentioned in papers~\citep{LIAMNE, MVE}, yet have not been discussed so far. As such, we also introduce a representation taxonomy to categorize the different kinds of representations, allowing for a deeper comparison between embedding models.

\subsection{Representation Taxonomy}\label{RepTaxonomy}

The task of network representation (or 
embedding) learning is to map the network information onto lower-dimensional latent variables $\zz\in\R^d$, where $d\ll N$~\citep{MEGAN,NetEmbTaxonomies}. Similarly, the space $\R^d$ is called the latent space or embedding space~\citep{ CuiEmbReview,zooGuide}. Prior works consider graph, edge and node embeddings. Here, we will focus only on node embeddings, which are the most commonly used~\citep{zooGuide}. 

The embedding model is a mapping $f:V \to \R^d$, so that $f(v_i)=\zz_i$ for all $v_i\in V$, where $\zz_i$ is the representation corresponding to node $v_i$. ~\cite{zooGuide} gave five criteria for an efficient embedding method for monoplex networks: \textbf{(i)} Adaptability: it should be applicable to a variety of different types of networks: (un)directed, (un)weighted, and so on; \textbf{(ii)} Scalability: the application to bigger networks should be computationally feasible; \textbf{(iii)} Topology awareness: it should capture structural differences in the network, for instance, due to strongly or weakly connected nodes; \textbf{(iv)} Low dimensionality: it should encode the information into a lower-dimensional space; and finally \textbf{(v)} Continuity: the embedding space should be continuous.

However, the extension of these five criteria to multiplex networks is not so straightforward. Node representations in multiplex networks need to incorporate information from the whole network, across layers (\textit{collaboration}), and to preserve the information of each layer individually (\textit{preservation})~\citep{MVN2Vec}. Types of representations can then be categorized into three distinct groups.

First, unique embedding, or one-space models~\citep{MVN2Vec}, or joint representation learning~\citep{LIAMNE} producea single  
representation per node~\citep{mpx2vec,RMNE}. This representation also holds for all replica nodes in other layers. Mathematically, we write that $f(v^{[\alpha]}_i;\mathcal{G})=\zz_i$ for all $v^{[\alpha]}_i\in V_{\alpha}$ and for all $\alpha \in Y$.
 
Second, in enriched embedding, or coordinated representation learning~\citep{LIAMNE},each replica node has its own representation~\citep{MANE, MNE, LIAMNE}. Nevertheless, these representations include information from the other layers and, as such, are not merely a single-layer embedding. This can be formalized as follows: $f(v^{[\alpha]}_i;\mathcal{G})=\zz_i^{[\alpha]}$ for all $v^{[\alpha]}_i\in V_{\alpha}$ and for all $\alpha \in Y$. 

Finally, for numerous embeddings, any node in any layer is associated with multiple embedding vectors according to its role in the network~\citep{MELL}, \eg, with a source and target node representation. Formally, this becomes $f(v_i^{[\alpha]};\mathcal{G})=\{\zz^{[\alpha]}_{i,1},...,\zz^{[\alpha]}_{i,k}\}$, where $k$ is the number of embeddings per node.

Some embedding models on multiplex networks also use so-called context embeddings~\citep{MVN2Vec, refinedMDeepWalk}. These were inspired by the word2vec model from~\cite{SkipGramFirst}, which represents words as contexts for other words within a sentence, and were later applied to embedding learning in networks~\citep{DeepWalk}. However, they do not fall into the scope of our review, as context embeddings are not used in the subsequent prediction tasks, but discarded after the embedding learning step. These types of embeddings are found in various models, including random walk and optimization-based models, as described in Section~\ref{taxonomy}.

\begin{figure}[h!]
    \centering
    \includegraphics[width=0.8\linewidth]{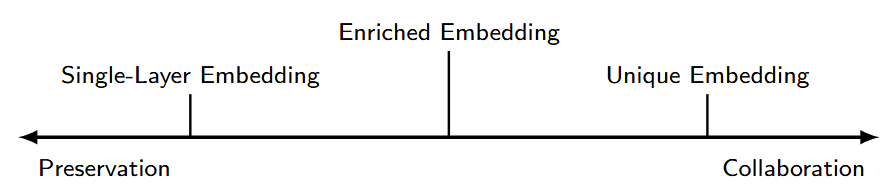}
    \caption{
    Single-layer embeddings, enriched embeddings, and unique embeddings fall onto the same spectrum of preservation and collaboration~\citep{MVN2Vec}.}
    \label{fig:collabpreserv}
\end{figure}

Unique embeddings implement a strong collaboration~\citep{MVN2Vec}, while single-layer embeddings are reflective of preservation without collaboration. Enriched embeddings might achieve a combination of collaboration and preservation. This spectrum is represented in Figure \ref{fig:collabpreserv}. On the one hand, unique embeddings may not preserve enough layer-specific information to perform well on the task of layer-specific link prediction. As an example, without loss of generality, consider the binary predictor $\mu:\R^d\times\R^d\to \{0,1\}$ that indicates the absence or the presence of an edge given the embeddings of the corresponding nodes.  Then, resorting to our notation, $\mu(f(v_i^{[\alpha]};\mathcal{G}), f(v_j^{[\alpha]};\mathcal{G})) = \mu(\zz_i,\zz_j)$. This means that such a predictor on unique embeddings would add or subtract the edge to every layer in the network. On the other hand, enriched embeddings are designed to overcome this shortcoming. Furthermore, considering the widespread use of symmetric prediction methods (see Section \ref{predictionMethod}), numerous embeddings--specifically target and source node representations--might considerably improve edge predictions for directed networks, as they reflect asymmetry. Moreover, numerous embeddings allow for better role-specific representations and thus increased topology awareness (criterion \textbf{(iii)}), though the number of embeddings directly impacts low-dimensionality (criterion \textbf{(iv)}).

All in all, the choice of embedding representation involves balancing the amount of preservation and collaboration required to tackle the downstream task, and should be carefully considered depending on the type of embedding techniques, which we review in the next section. 

\subsection{Method Taxonomy}\label{taxonomy}

Many taxonomies have been proposed for embedding techniques in monoplex networks, as evidenced by the number of available surveys~\citep{CuiEmbReview,  NetEmbTaxonomies, NetRepLASurvey, zooGuide}, though \textit{not} in multiplex networks as in our review. 

For instance, \cite{NetEmbTaxonomies} gives a two-level taxonomy. The first one splits methods into embeddings of dynamic and static monoplex networks. The second level classifies the static methods similarly to~\cite{CuiEmbReview} into embeddings with structural information, and embeddings with additional information (\eg, node attributes). Yet, as the scope of this paper is on multiplex networks without additional information, this taxonomy is not relevant in our case. 

~\cite{NetRepLASurvey} make a distinction between unsupervised and supervised models for embedding learning in monoplex networks. While this classification makes sense, we believe the taxonomy introduced by~\cite{zooGuide} to be more adequate, as it categorizes models into groups of common mathematical approaches instead of learning paradigms. The naive extension of these groups to multiplex networks is straightforward, yet it would neglect a whole class of methods based on an aggregation approach across layers. 

The aforementioned review on  clustering~\citep{MultiplexClusterReview} mentions traditional clustering approaches on networks, which do not use embeddings, generalized to multiplex networks, based on two main procedures. The first one involves the transformation of the multiplex network into a single-layer network, followed by the application of a  clustering method for monoplex networks. The second procedure consists in individually applying to each layer a clustering method for monoplex networks and then combining the results.

~\cite{WhereFuseInformation} introduce a taxonomy of methods for embedding learning named ``fusion'' restricted to graph neural networks (GNNs). Their taxonomy identifies four types of fusion: graph-level, GNN-level, embedding-level, and prediction-level fusions. However, it is important to note that both prediction-level and GNN-level fusion are not applicable to the context of network representation. Indeed,
all models inherently combine information, rendering GNN-level fusion unnecessary, while prediction-level fusion does not influence the embeddings.

We propose a method taxonomy with two levels, similarly to~\cite{zooGuide}.
We first make a high-level distinction 
between shallow and deep methods. In short, deep methods have hidden layers, shallow methods do not. As such, it is straightforward to see that neural network-based methods (\eg graph convolution neural networks, or GCNs) are deep methods, and all other methods are shallow methods. The second lower-level level of taxonomy considers random walk, optimization, matrix factorization, and aggregation-based methods, the latter being more isolated from all other former groups. Over the next few sections, we will describe the characteristics of each group. We illustrate the distinction between representation and method taxonomies in Table~\ref{tab:MethodsVSRepresentationTable}.

\begin{table}[ht]
    \centering
    
    \caption{Representation and method taxonomies on the literature on embedding learning in multiplex networks for link prediction. The rows correspond to the type of representation, whereas the columns are organized by the technique
    used in the embedding model.}
    \label{tab:MethodsVSRepresentationTable}
    \footnotesize
    \begin{tabularx}{\linewidth}{lccccc}
      \toprule
         &
        \multicolumn{5}{c}{Embedding Method}\\
      \cmidrule(lr){2-6}
        &
        \multicolumn{3}{c}{Shallow} &
        \multicolumn{1}{c}{Deep}\\
      \cmidrule(lr){2-4}\cmidrule(lr){5-5}
       \multicolumn{1}{l}{Embedding}  &
        \multicolumn{1}{c}{Matrix} &
        \multicolumn{1}{c}{Random} &
        \multicolumn{1}{c}{} &
        \multicolumn{1}{c}{Neural} &
        \multicolumn{1}{c}{} \\

        \multicolumn{1}{l}{Representation}  &
        \multicolumn{1}{c}{Factorization} &
        \multicolumn{1}{c}{Walk} &
        \multicolumn{1}{c}{Optimization} &
        \multicolumn{1}{c}{Networks} &
        \multicolumn{1}{c}{Aggregation} \\
        
      \cmidrule(r){1-1}\cmidrule(lr){2-2}\cmidrule(lr){3-3}\cmidrule(lr){4-4}
        \cmidrule(lr){5-5}\cmidrule(lr){6-6}
        \multirow{9}{*}{\shortstack[l]{Unique\\Embedding}}   &   & RMNE &  & MGAT & PMNE\\
         &    & MDeepWalk & &  MultiplexSAGE  & \\
          &   & FFME & & DGMI & \\
          &  & mpx2vec & & mGCN & \\
          &  & MultiVERSE & & VANE & \\
        &   & MANE+ & & MEGAN & \\
         &  & MVN2VEC & &  & \\
          &  & MHME & &  & \\
            &  & Multi-node2vec &  &  & \\
           &  &  &  &  & \\
        
        \multirow{4}{*}{\shortstack[l]{Enriched\\Embedding}}   & MANE & MNE & MTNE & CGNN & MNI\\
         &  & NANE & MVE & LIAMNE & \\
           &  & RWM & MUNEM & MNI-DGI & \\
          &  & OhmNet & &  & \\
          &  &  & &  & \\
        
        \multirow{2}{*}{\shortstack[l]{Numerous\\Embedding}}    &    &  & MELL &  &    \\
         &  &  &  &  & \\
      \bottomrule
    \end{tabularx}
\end{table}

\subsubsection{Aggregation Methods}\label{aggregation}

\begin{figure}[!ht]
    \includegraphics[width=\textwidth]{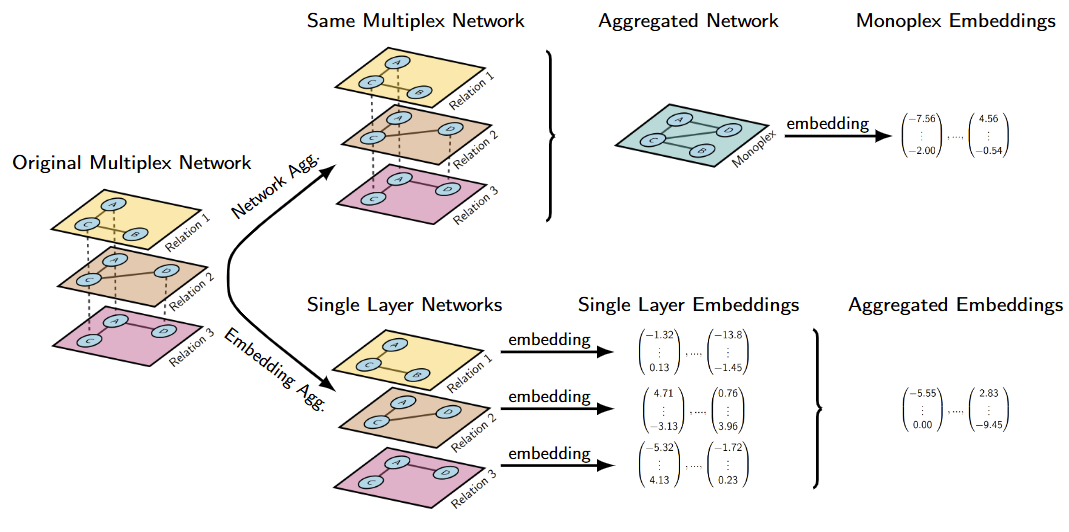}
    \caption{Difference between network-level and embedding-level aggregations. The aggregation step is depicted by the braces.}
    \label{fig:AggregationFigures}
\end{figure}

As the name suggests, the aggregation, also called fusion~\citep{WhereFuseInformation}, method merges or fuses information together in the multiplex network. We can distinguish between a network-level and an embedding-level information aggregation. The latter is sometimes referred to as \textit{results aggregation}~\citep{PMNE}. A visual depiction of both aggregation approaches is given in Figure~\ref{fig:AggregationFigures}.

First, in network aggregation, the information of the multiplex network is fused down to produce a simple network onto which the embedding method is then applied. In other words, the goal is to construct a graph $\tilde{G}=(V,\tilde{E})$ 
from $\mathcal{G}$ by some method $\text{Agg}$ with $\tilde{\aA}=\text{Agg}(\aA_1,\dots,\aA_{M})$ the corresponding adjacency matrix for the aggregated network. The difficulty lies in fusing the multiplex into a single network, while retaining as much information on the diverse node relations as possible. Different types of network-level aggregation have been introduced (usually in the realm of complex system physics), for instance, %among others: 
average network, projected monoplex network, and overlay network~\citep{DimRed&SpecPropofMulNet}. One of the PMNE model~\citep{PMNE} variants combines the layers of the multiplex network into a weighted network by summing the edge weights; formally: $\text{Agg}(\aA_1,\ldots,\aA_M) = \sum_{\alpha\in Y} \aA_\alpha$.

In any case, aggregating the layers into a single network implies information loss, as it somewhat neglects the preservation objective from Section~\ref{RepTaxonomy}. This information loss was quantified and leveraged by~\cite{de_domenico_structural_2015} to minimize loss during aggregation.

On the other hand, the embedding-level aggregation first applies an embedding method for monoplex networks on each layer to obtain one representation for each replica node, and then aggregates the representations of an entity across layers to produce a single representation  
for all replica nodes of a single entity. More formally, the goal is to construct a latent vector $\tilde{\zz}_i\in\R^d$ to node $v_i\in V$ from embeddings $\zz_i^{[1]},\ldots,\zz_i^{[M]}$ associated with replica nodes $v_i^{[1]}, \dots, v_i^{[M]}$ by some method $\text{Agg}$, with $\tilde{\zz}_i=\text{Agg}(\zz_i^{[1]},\ldots,\zz_i^{[M]})$.  
An example of an aggregation function would be the mean across layers of replica representations, as done for a baseline method in~\cite{MultiVERSE}. Another variant of the PMNE model~\citep{PMNE} directly concatenates the layer-specific representations, that is, $\text{Agg}(\zz_i^{[1]},...,\zz_i^{[M]}) = \text{concat}_{\alpha\in Y} \zz_i^{[\alpha]}\in\R^{Md}$. A more sophisticated approach in the MNI model~\citep{MNI} aggregates the embeddings for each layer by maximizing the mutual information between the latent vectors. Depending on the approach, embedding-level fusion can allow for some collaboration, whilst preserving more layer-specific information.
 
\cite{PMNE} and \cite{WhereFuseInformation} compared the performance of models when using network-level aggregation, embedding-level aggregation, and a version not based on aggregation, allowing for more collaboration. They concluded that the aggregation methods performed the worst for link prediction. This might explain why, while aggregation is a straightforward option, it is rarely used. Some models still use embedding-level aggregation after having implemented a collaborative model to retrieve unique embeddings~\citep{MVN2Vec, MANE+}, though it is rare. However, aggregation is still prominently used with monoplex-network-specific embedding methods as baseline methods~\citep{PMNE}. Indeed, embedding models for multiplex networks are often compared to LINE~\citep{LINE}, DeepWalk~\citep{DeepWalk}, or struc2vec~\citep{Struc2Vec} combined with either network-level~\citep{MEGAN,FFME&MHME, LIAMNE} or embedding-level~\citep{MUNEM, MultiVERSE} fusion.

\subsubsection{Matrix Factorization}

Matrix factorization models take advantage of the fact that the (potentially weighted) adjacency matrix $\aA$ holds all the information about the graph~\citep{zooGuide}. Therefore, these methods apply classical matrix factorization techniques from linear algebra to obtain representation vectors. In particular, those methods often rely on the definition of the Laplacian $\lL$ of a network, defined by the difference of the degree and the adjacency matrices of the network, \ie $\lL=\dD-\aA$, where the degree matrix $\dD$ is a diagonal matrix containing the degree of each of the $N$ nodes. Moreover, we also define the normalized Laplacian matrix $\widehat{\lL} = \dD^{-1/2}\lL\dD^{-1/2}$.

A comprehensive list of matrix factorization approaches developed for monoplex networks can be found in~\cite{zooGuide}. A notable example is the Laplacian Eigenmaps~\citep{LaplacianEigenmaps}, where embedding vectors preserve the closeness of the nodes in the network with respect to their edge weights. Laplacian Eigenmaps are obtained by maximizing a trace involving the Laplacian of the network and the matrix of concatenated embedding vectors $\zZ\in\R^{N\times d}$: $\max_{\zZ} \text{tr}\left(\zZ^\intercal \lL \zZ\right)$, under the constraint $\zZ^\intercal \dD\zZ=\iI_d$.\footnote{The trace $\text{tr}(M)$ of a matrix $M$ takes the sum of the diagonal elements of the matrix.} The constraint allows the authors to remove an arbitrary scaling factor in the embeddings.

In MANE~\citep{MANE}, this trace maximization approach is extended to the multiplex case by computing enriched embeddings. To this end, the authors split the objective into two components: (1) the intralayer objective, which is the straightforward extension of the monoplex formulation to each layer: $\max_{\zZ_\alpha} \text{tr}(\zZ_\alpha^\intercal\widehat{\lL}_\alpha \zZ_\alpha)\text{ under the constraint }
\zZ_\alpha^\intercal \zZ_\alpha=\iI$ for all layer $\alpha \in Y$; and (2) the interlayer objective, which is: 
\begin{equation*}
    \min_{\zZ_\alpha, \zZ_\beta, \kK_{\alpha\beta}}\sum_{\alpha,\beta=1}^{M}{\|\iI_N -\zZ_\alpha^\intercal \kK_{\alpha\beta}\zZ_{\beta}\|_F^2}, \text{ such that }\zZ_\alpha^\intercal \zZ_\alpha=\iI_N, \text{ for all layers $\alpha\in Y$,}
\end{equation*}
where $\kK_{\alpha\beta}\in\R^{n\times n}$ is the interlayer adjacency matrix between layers $\alpha$ and $\beta$. Together, these objectives are merged into the following trace maximization problem:
\begin{equation*}
    \max_{\zZ_\alpha} \text{tr}\left(\zZ_\alpha^\intercal \left(\widehat{\lL}_\alpha+\lambda\sum_{\beta\in Y}{\zZ_{\beta}\zZ_{\beta}^\intercal}\right)\zZ_\alpha\right),\text{ such that }\zZ_\alpha^\intercal \zZ_\alpha=\iI_N\text{, }\forall\alpha\in Y.
\end{equation*}
The formulation above is the multiplex case of the more general multilayer network description in~\cite{MANE}. MANE has so far been the only published work to use matrix factorization for generating node embeddings on multiplex networks. Nevertheless, random walk approaches, which will be discussed in the next section, appear to be closely related to matrix factorization, as some researchers have noticed~\citep{RWAndMF}.

\subsubsection{Random Walk-Based Methods}\label{random-walk}

\begin{figure}[!ht]
    \begin{subfigure}[b]{0.24\textwidth}
         \centering
         \includegraphics[width=\textwidth]{Fig2.png}
         \caption{}
         \label{fig:multiplexForRW}
     \end{subfigure}
     \hfill
     \begin{subfigure}[b]{0.24\textwidth}
         \centering
         \includegraphics[width=.97\textwidth]{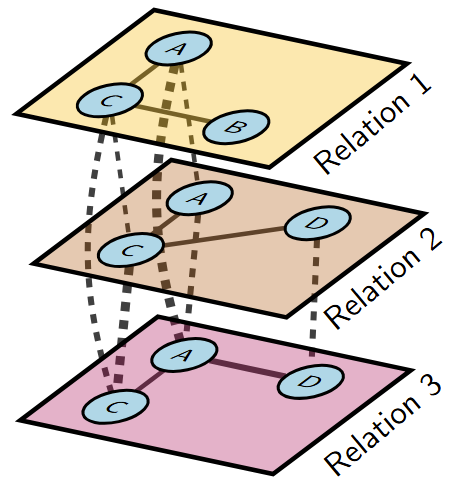}
         \caption{}\label{fig:transitionProbabilities}
     \end{subfigure}
     \hfill
     \begin{subfigure}[b]{0.24\textwidth}
         \centering
         \includegraphics[width=\textwidth]{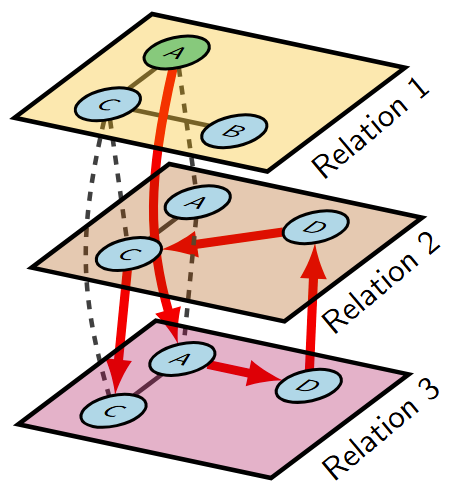}
         \caption{}
         \label{fig:BiasedRW}
     \end{subfigure}
     \hfill
     \begin{subfigure}[b]{0.24\textwidth}
         \centering
         \includegraphics[width=\textwidth]{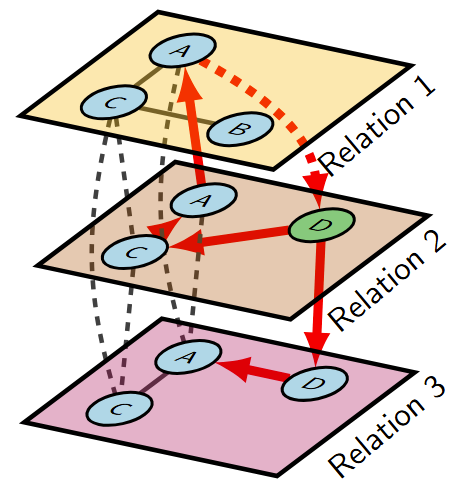}
         \caption{}
         \label{fig:RWR}
     \end{subfigure}
     \hfill
    \caption{Figure \ref{fig:multiplexForRW} shows the original multiplex network. Figure \ref{fig:transitionProbabilities} displays the transition probabilities for the random walk: the thicker the line, the higher the probability of walking this edge.
    Figure \ref{fig:BiasedRW} depicts a random walk using the most likely node transitions, where the green node is the seed of the random walk, and the red arrows the path taken. Figure \ref{fig:RWR} illustrates a random walk \textit{with restart} on a multiplex network, where the dashed red arrow indicates the teleportation back to the seed node.}
    \label{fig:twographs_randomwalk}
        
\end{figure}

A random walk on a graph is a sequence of points obtained by starting at an initial node, randomly selecting an edge to traverse to the next node, and then iterating this procedure~\citep{RandomWalksOnGraphs}. Typically, edges are selected with a probability proportional to the edge weight, which allows the random walker to explore the network structure. The element $[\pP]_{ij}$ of the transition matrix $\pP=\dD^{-1}\aA$ gives the probability that a random walker will move from node $v_i$ to $v_j$, for $\dD$ and $\aA$ respectively the degree and adjacency matrix of a monoplex network. Formally, we write this as $p(v_i,v_j)=[\pP]_{ij}=w(v_i,v_j)/\deg(v_i)$. The position update from step $n$ to step $n+1$ of the random walker is achieved by computing $\mm_{n+1} = \pP^\intercal \mm_n$, where $\mm_n$ and $\mm_{n+1}$ are probability vectors in $\R^N$, where $N$ is the number of nodes in the graph. This process, known as the standard random walk and its variants, is at the heart of this section. A visualization of the two most central random walk methods, random walk and random walk with restart--where the random walker is allowed to teleport back to the starting node (referred to as the \textit{seed node}~\citep{MultiXrank})--is given in Figure~\ref{fig:twographs_randomwalk}.

Random walk-based methods for embedding learning on monoplex networks began with DeepWalk~\citep{DeepWalk}, which was inspired by the natural language processing model word2vec~\citep{SkipGramFirst}. Since then, many researchers have worked on advancing the approach, introducing prominent variants such as node2vec~\citep{node2vec} or struct2vec~\citep{Struc2Vec}. node2vec~\citep{node2vec} adapted the DeepWalk model by using a biased second-order random walk--meaning that the walker remembers the last step taken--and using the Skip-Gram objective~\citep{SkipGramFirst} combined with Alias table sampling~\citep{AliasTableMethod} for optimization. The Skip-Gram objective takes in a sampled sequence of nodes, and maximizes the likelihood for two nodes close to each other in the sequence. Formally, this is written as
\begin{equation}\label{eq:SkipGram}
    \max\frac1{T}\sum_{i=1}^{T}\sum_{-c\leq j\leq c, j\neq i}\log p(v_{i+j}|v_i),
\end{equation}
where $T$ is the size of the node sequence $S:=\{v_1,v_2,...,v_T\}$, $\{v_{i-c},...,v_{i-1}, v_{i+1},...,v_{i+c}\} \cap S$ is the context window for node $v_i$, and $c\in\N$ is a constant and a hyperparameter of the objective. The Alias table sampling method is the random walk adaptation \citep{node2vec} used to sample the node sequences.

Another approach on monoplex networks, struc2vec~\citep{Struc2Vec}, focuses on retrieving structurally informed embeddings. To this end, the authors construct a weighted multiplex network, where the $k$-th layer represents $k$-step connections, with the first layer being the original network. A $k$-step connection exists between two nodes if and only if one can go from one node to the other by walking exactly along $k$ edges in the network. Then, the random walker starts at a node in the first layer and traverses the multiplex, where it is motivated to travel upwards to capture only the most structurally similar nodes. The path through the multiplex represents the sequence of nodes fed into the Skip-Gram objective function. These methods form a group of random walk methods, which we call \textit{random walk sampling} procedures, as the goal of the random walk is to sample a sequence of nodes that will be fed to an objective function.

Unlike \cite{zooGuide}, we also consider random walk variants in which embedding learning is performed on the probability states $\mm^*$, for instance, in VERSE~\citep{VERSE}. VERSE uses the random walk with restart process defined in step $n>0$ by:
\begin{equation}\label{RandomWalkWithRestart}
    \mm_{n+1} = (1-r)\pP^\intercal \mm_n + r\mm_0,
\end{equation}
where $r\in(0,1)$ is the probability that the random walker will be teleported back to the seed node at each step. In the case where $r=0$, we retrieve the standard random walk. However, for $r=1$, the random walker does not move from his initial position. The parameter $r$ thereby determines how local the walk will be. This process converges to a steady state $\mm^*$ as $n\to\infty$~\citep{RWRConvergence}. If $\mm_0$ is a unit vector, with $[\mm_0]_{i}=1$ and the other elements being $0$, then $[\mm^*]_j$ can be understood as a probability value for the proximity of node $v_j$ to the unique seed node $v_i$. VERSE leverages this to construct embeddings by minimizing the cross-entropy between the resulting probability values from the random walk with restart process, and an estimator. We refer to these methods as \textit{random walk limiting} procedures.

\cite{ABroaderPicture} proposed a general structure for random walk embedding approaches on monoplex networks. Three major components were identified as common to all random walk-based methods: the random walk process; the similarity function; and finally, the embedding algorithm. While these components have only been applied to monoplex networks and random walk sampling procedures, we will demonstrate that this framework can be extended to random walk-based methods for multiplex networks, including random walk limiting procedures. We use these components to explain the similarities and differences between the models.

\paragraph{Random walk process.} \cite{ABroaderPicture} classify the random walk processes of monoplex models into three categories: standard random walk, biased random walk, and PageRank~\citep{PageRank}. A biased random walk is a random walk with modified transition probabilities to prioritize certain transitions over others. The PageRank method teleports the random walker to a new (not necessarily connected) node with some probability at each iteration. In the multiplex context, we found only models using the random walk with restart (also known as rooted PageRank) or biased random walk algorithms.

The biased random walk is a truncated random walk with a fixed number of steps and transition probabilities that are no longer proportional to the edge weights. This method is used for node sequence sampling, which makes it a random walk sampling procedure. The general formula for the transition probability of the walker going from node $i$ in layer $\alpha$ to node $j$ in layer $\beta$ is given by~\cite{mpx2vec}:
\begin{equation}\label{generalRandomWalk}
    p(v_i^{[\alpha]},v_j^{[\beta]})=\begin{cases}
        g_s(v_i^{[\alpha]},v_j^{[\beta]}) \quad\text{if } \alpha = \beta\\
        g_d(v_i^{[\alpha]},v_j^{[\beta]}) \quad\text{if } i = j\\
        0,\quad\text{else}
    \end{cases}
\end{equation}
where $g_s$ describes the intralayer transition rule and $g_d$ describes the interlayer transition rule. Diverse interlayer transition rules exist to achieve collaboration. 

For instance, RMNE~\citep{RMNE} favors interlayer transitions according to the structural similarity of the nodes. The walker can travel to neighboring nodes and structurally similar nodes, such as those with similar centrality and outgoing degree. For the interlayer transition, NANE~\citep{NANE} emphasizes using structurally similar replica nodes to reduce the ``noise" in the embeddings from layers that differ significantly from the layer under consideration (named the \textit{target layer}). They achieve this by comparing the vectors of the weighted averages of $k$-step adjacency matrices through the cosine similarity. Another approach for interlayer transitions is used in MDeepWalk~\citep{refinedMDeepWalk}, which extends the monoplex DeepWalk method to a multiplex scenario, where $g_d$ equals to the Jaccard coefficient~\citep{JaccardScore} on the intralayer neighbors of 
replica nodes $v_i^{[\alpha]}$ and $v_i^{[\beta]}$.
\begin{equation*}
    g_d(v_i^{[\alpha]}, v_i^{[\beta]}) = \frac{\#\left(\mathcal{N}(v_i^{[\alpha]})\cap\mathcal{N}(v_i^{[\beta]})\right)}{\#\left(\mathcal{N}(v_i^{[\alpha]})\cup\mathcal{N}(v_i^{[\beta]})\right)}\:,
\end{equation*}
where $\mathcal{N}(v_i^{[\alpha]})$ and $\mathcal{N}(v_i^{[\beta]})$ are the intralayer neighborhoods of $v_i^{[\alpha]}$ on layer $\alpha$ and $v_i^{[\beta]}$ on layer $\beta$, respectively. As a result, the random walker is more likely to move to replica nodes with similar neighborhoods. Similarly, Multi-node2vec~\citep{Multi-node2vec} extends the node2vec model to the even more general multilayer networks~\citep{MultilayerReview2014}, by sampling nodes in the network as in node2vec with a second-order biased random walk. Multiplex networks are a special case of multilayer networks where only interlayer connections between replica nodes are allowed.

A different approach is taken by mpx2vec~\citep{mpx2vec}, which constructs \textit{a priori} a node importance matrix $\mathbf{J}\in\R^{N\times M}$, with rows summing up to $1$, and allocates a weight to each replica node. This matrix is first randomly initialized and then updated through a PageRank-inspired algorithm until convergence. The transition probabilities are then proportional to the product of those weights and to the distance $\tau$ between replica nodes on the same layer $\alpha$: $p(v_i^{[\alpha]},v_j^{[\beta]}) = \tau(v_i^{[\alpha]},v_j^{[\alpha]}) \cdot [\mathbf{J}]_{i\beta}$. 

\cite{FFME&MHME} criticize the inherent bias of the random walk towards highly connected nodes. To counteract this bias, the authors propose two models, FFME and MHME. The interlayer transition rules in both models depend on the neighbor's partition coefficient (NPC), which they defined as:
\begin{equation*}
   \text{NPC}(v_i) = \frac{2}{\#\tilde{Y}(\#\tilde{Y}-1)}\sum_{\alpha< \beta \leq M}\frac{\#\left(\mathcal{N}(v_i^{[\alpha]})\cap\mathcal{N}(v_i^{[\beta]})\right)}{\#\left(\mathcal{N}(v_i^{[\alpha]})\right)},
\end{equation*}
where $\tilde{Y}$ is the set of layers $\alpha$ in which the node $v_i^{[\alpha]}$ is not isolated, \ie of total degree $0$. The NPC is used to ensure that interlayer jumps are primarily made between replica nodes with different neighborhoods. The specific models then construct a multiplex network random walk that is designed with either a Metropolis-Hastings strategy (leading to the MHME model), or forest fire sampling~\citep{ForestFireSampling} method (FFME model). The Metropolis-Hastings strategy modifies the random walk process to ensure that the resulting steady state has a uniform distribution over the nodes, demonstrating that this method does not favor highly connected nodes. On the other hand, the forest fire sampling method is able to capture the modularity of the network. Their experiments show that, on the task of link prediction, the MHME model generally outperforms the FFME model.

Unlike the previous methods, some methods independently perform a random walk on each layer. For instance,~\cite{RWM} propose two variants of their model RWM: one with a biased random walk, and the other with a truncated random walk with restart. As the walkers traverse their respective network layers, they also influence the walkers in the other layers according to how similar their local structures are. This influence is captured in a relevance matrix that is updated with each step. This is how they achieve the collaboration in the resulting embedding. 

In contrast to the aforementioned models, RWM requires a precomputed interlayer transition rule. Other models also fall into this category, applying the random walk on each layer separately and achieving collaboration in the optimization stage of the algorithm. This class includes the models OhmNet~\citep{OhmNet}, MVN2Vec~\citep{MVN2Vec}, MNE~\citep{MNE}, and MANE+~\citep{MANE+}--not to be confused with MANE~\citep{MANE} from the matrix factorization section. All of these truncated models require the number of steps of the truncated random walk to be set beforehand; a typical choice is a length of 10 steps.

The MultiVERSE~\citep{MultiVERSE} extends the VERSE algorithm on monoplex networks to multiplex networks by using a random walk with restart method and is, therefore, a random walk limiting procedure. MultiVERSE introduces interlayer transition probabilities $(\lambda_{\alpha})_{\alpha \in Y}$ as hyperparameters, with $\sum_{\alpha\in Y}{\lambda_{\alpha}}=1$. At each node in layer $\alpha$, the walker moves to a replica node in layer $\beta \neq \alpha$ with probability $\lambda_{\beta}$ or travels to an intralayer neighboring node layer $\alpha$ with probability $\lambda_{\alpha}$. In other words, the probability of walking from node $i$ on layer $\alpha$ to node $j$ on layer $\beta$ is: 
\begin{equation*}
    p(v_i^{[\alpha]},v_j^{[\beta]}) =\begin{cases}
        \lambda_\alpha\cdot p^{[\alpha]}
        (v_i^{[\alpha]},v_j^{[\alpha]}) \text{ if }\alpha=\beta\:,\\
        \lambda_\beta \text{ if }i=j,
    \end{cases}
\end{equation*}
where $p^{[\alpha]}(v_i^{[\alpha]},v_j^{[\alpha]}) = [\dD_\alpha^{-1} \aA_\alpha]_{ij}$ is the $i$-th row and $j$-th column of the transition matrix of layer $\alpha$. For this model, the transition hyperparameters $(\lambda_\alpha)_\alpha$ must be specified beforehand. In their paper, they simply set $\lambda_\alpha = \delta$ and $\lambda_\beta = (1-\delta)/(M-1)$, which yields the following supra-transition transition matrix:
\begin{equation*}
    \pPP=\begin{pmatrix}
        \delta \pP_1 & \frac{1-\delta}{M-1}\iI_{N} & \hdots & \frac{1-\delta}{M-1}\iI_{N}\\
        \frac{1-\delta}{M-1}\iI_{N} & \delta \pP_2 & \ddots &\vdots\\
        \vdots & \ddots & \ddots & \frac{1-\delta}{M-1}\iI_{N}\\
        \frac{1-\delta}{M-1}\iI_{N} & \hdots & \frac{1-\delta}{M-1}\iI_{N} & \delta \pP_M
    \end{pmatrix}.
\end{equation*}

\paragraph{Similarity function.} The node similarity function $\varphi: V \times V \to [0,1]$ measures the topological similarity between pairs of nodes~\citep{ABroaderPicture}. A value closer to $1$ indicates a higher similarity between the two nodes, while a value closer to $0$ suggests lower similarity. These similarity functions are integral to the optimization objective and play a key role in quantifying the relationships between nodes. All previously mentioned models~\citep{mpx2vec,refinedMDeepWalk,FFME&MHME,MultiVERSE,RMNE,RWM} use the sigmoid as their similarity function 
$\varphi(v_i,v_j) =\sigma( \cc_i^\intercal \zz_j)\:,$ where $\sigma(x)=1/(1+\exp(-x))$ and $\cc_i$ is the context vector representation of $v_i$, and $\zz_j$ is the embedding of node $v_j$. \cite{ABroaderPicture} sort this similarity function into a class that they refer to as pointwise mutual information-based (PMI) functions. They examined the effect of using another type of similarity function called autocovariance~\citep{Autocovariance}, which they report outperformed PMI functions on link prediction tasks. Their analysis has never been extended to multiplex networks, and may be an interesting route for further research.

\paragraph{Embedding algorithm.} Finally, regarding the embedding algorithm, \cite{ABroaderPicture} distinguishes between matrix factorization and sampling techniques. We are only aware of negative sampling techniques being used on multiplex networks~\citep{MNE,refinedMDeepWalk,  RMNE}. The idea behind negative sampling is to randomly select several elements outside the original dataset for each element within the dataset, aiming to improve the model's performance while maintaining computational efficiency. Most models work analogously to node2vec, and feed the node sequences to the Skip-Gram objective~\citep{SkipGramFirst} (see Equation~\eqref{eq:SkipGram}), which is then optimized using stochastic gradient descent~\citep{mpx2vec, MNE, NANE, refinedMDeepWalk, Multi-node2vec}. Since many of the real-world networks have a large number of nodes, it is intractable to optimize over all possible edges. 

The Skip-gram algorithm with negative sampling~\citep{SkipGram} achieves this by corrupting some of the existing edges and ensuring that these negative edges do not belong to the set of edges sampled during the random walk. This scheme is used to approximate the log of the conditional probability inside Equation~\eqref{eq:SkipGram}, as computed by the $\text{Softmax}$, by:
\begin{equation*}
    \log p(v_{i+j}|v_i) = \log\sigma(\cc_{i+j}^\intercal \zz_{i}) + b\cdot\mathbb{E}_{\cc\sim p(\cc)}\left[\log\sigma(-\cc^\intercal \zz_i)\right]\:,
\end{equation*}
where the $\cc$'s are context embeddings, $p(\cc)$ is the distribution from which the fake edges are sampled, and $b$ is the number of negative samples. Another method to reduce the computational cost of the Softmax in the Skip-Gram is noise contrastive estimation (NCE)~\citep{NCE}, although it has not been used in multiplex network embedding. The NCE can be shown to approximately maximize the conditional log likelihood in the Skip-Gram objective, whereas the Skip-Gram with negative sampling is only a simplified version of NCE, based solely on heuristics~\citep{SkipGram}.

In models with layer-specific random walks, collaboration is achieved through the embedding algorithm. MNE~\citep{MNE} facilitates this collaboration by replacing the enriched node embeddings with the sum of a shared representation across layers, combined with a transformation of the layer-specific embedding. The transformation function is learned during the optimization process. In OhmNet~\citep{OhmNet}, the layers are arranged hierarchically, reflecting the biological tissues considered. The algorithm is designed to convey hierarchical information as described by a tree, by ensuring that the node embeddings of replica nodes from parent and child networks are similar, as measured by the $\ell_2$-distance. Finally, in MVN2VEC-REG~\citep{MVN2Vec}, layers are not hierarchically organized, but the representations specific to each layer are regularized by the average representation of replica nodes, thus, ensuring that the enriched embeddings are close to their mean value across replicas.

A very different approach is considered in MANE+~\citep{MANE+}, where, in addition to the Skip-Gram objective for each layer, the model uses a Skip-Gram objective with interlayer connections. The goal is to capture what they refer to as first- and second-order collaboration in the multiplex network. These objective functions are summed to a final function.
 
The random walk approaches are popular for embedding learning on multiplex networks, which is reflected in their numerous publications. They are characteristically defined by their three-part structure, comprising a random process, a similarity function, and an embedding algorithm. This also separates them from the category of optimization methods, which is addressed below.

\subsubsection{Optimization Methods}\label{optimization}

\begin{figure}[!ht]
    \centering
    \includegraphics[width=\linewidth]{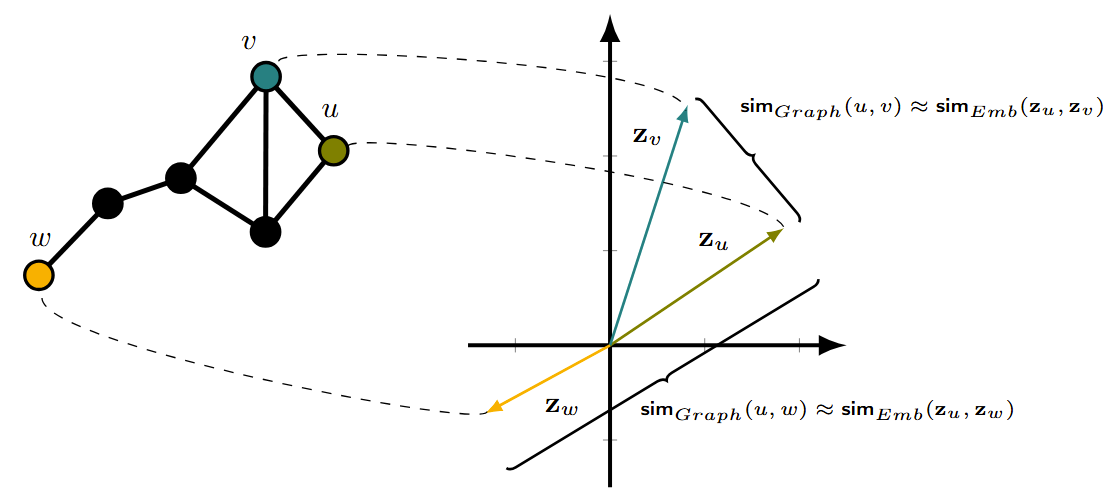}
    \caption{Optimization models mainly aim at preserving first-order and second-order proximity information, here on a monoplex network. This figure was inspired by Figure 2 in~\cite{zooGuide}.}
    \label{fig:Optimization}
\end{figure}

This section includes the remaining shallow methods that do not fit into the previous categories of matrix factorization and random walk-based approaches. The models in this section appear to share an underlying characteristic: each one is based on a direct pairwise comparison of node embeddings, aiming to preserve the similarities between the node pairs in the graph through some \textit{first-order} or \textit{second-order proximities}, as depicted in Figure~\ref{fig:Optimization}. The $k$-order proximity refers to the $k$-step connection between nodes $v_i$ and $v_j$, which can be measured by their transition probability $p^{(k)}(v_i,v_j) = \left[(\dD^{-1}\aA)^k\right]_{ij}$~\citep{NetEmbTaxonomies}.

A prominent example on monoplex networks is LINE~\citep{LINE}. This model focuses on preserving first-order and second-order proximities of a node by relying on prior-knowledge context embeddings $(\textbf{c}_i)_{v_i \in V}$ and by maximizing over an objective function on the embedding matrix $\zZ$ for the proximity of order $k$ on the monoplex network $\mathcal{G} = (V, E)$

\begin{equation}\label{eq:LINE_obj} 
    \text{prox}^{(k)}(\zZ; \mathcal{G}, (\textbf{c}_i)_{v_i \in V}) = -\sum_{(v_i,v_j)\in E}{p^{(k)}(v_i,v_j)\log p^{(k)}}(v_i,v_j; \zZ, (\textbf{c}_i)_{v_i \in V})\;.
\end{equation}
The predicted probability for the first-order proximity between nodes $v_i$ and $v_j$ with respective embeddings $\zz_i$ and $\zz_j$ is given by $p^{(1)}(v_i, v_j; \zZ) = \sigma\left(\zz_i^\intercal \zz_j\right)$, where $\sigma$ is the sigmoid function. On the other hand, the predicted probability of the second-order proximity uses the aforementioned context embedding  $\cc_i$ for node $v_i$ $p^{(2)}(v_i, v_j; \zZ, (\textbf{c}_i)_{v_i \in V}) = \sigma\left(\cc_i^\intercal \zz_j\right)$.

MTNE~\citep{MTNE} extends the LINE model to multiplex networks by computing enriched embeddings. The authors facilitate collaboration between the layers through two distinct approaches. The first collaboration approach is by computing so-called common embeddings, which are representations shared across layers. These embeddings are obtained by maximizing over an objective function similar to Equation~\eqref{eq:LINE_obj}, where the first-order and second-order proximities are defined as
\begin{eqnarray*}
    p^{(1)}(v_i^{[\alpha]}, v_j^{[\alpha]}; \zZ) & = &\sigma    \left(\big(\zz_i^0+\zz_i^{[\alpha]}\big)^\intercal \big(\zz_j^0+\zz_j^{[\alpha]}\big)\right)\\
    p^{(2)}(v_i^{[\alpha]}, v_j^{[\alpha]}; \zZ, (\textbf{c}^{\alpha}_i)_{v_i \in V}) & = & \sigma\left(\big(\cc_i^{\alpha}\big)^\intercal\big(\zz_j^0+\zz_j^{[\alpha]}\big)\right)\;,
\end{eqnarray*}
where $\zZ_0=[\zz_1^0, \ldots,\zz_N^0]^{^\intercal}$ is the common embedding matrix, and $\textbf{c}^{\alpha}_i$ is the context embedding associated with node $v_i$ on layer $\alpha$. Finally, the objective functions also include the Frobenius norms of each layer-specific concatenated embedding matrix ($(\zZ_\alpha)_\alpha$, $(\cC_\alpha)_\alpha$) as regularization terms. The second collaboration approach taken in MTNE is to use the so-called consensus embeddings. The authors use the LINE algorithm on each layer, with an additional regularization term through a consensus matrix $\kK\in\R^{N\times d}$ across layers such that the objective function includes $\sum_{\alpha\in Y}{\|\zZ_\alpha-\kK\|_F^2}+\lambda\|\kK\|_F^2\:,$ where $\lambda$ is a weighting parameter. This term forces the layer-specific embeddings to be more similar, thereby facilitating collaboration. On the task of link prediction,~\cite{MTNE} deduced from their experimental results that the common embedding approach mostly outperformed the consensus embedding one. They believed this was a result of the more explicit integration of layer-specific embeddings. 

Another approach on multiplex networks, MELL~\citep{MELL}, also relies on a LINE-like strategy. However, contrary to MTNE, MELL uses separate head (source node) and tail (target node) embeddings, respectively denoted $(\zZ_{\alpha,h})_{\alpha}, (\zZ_{\alpha,t})_{\alpha} \subset \R^{N\times d}$. The purpose of this distinction was to better predict links in directed networks. Moreover, MELL also introduces layer embeddings $\mathbf{R}\in\R^{M\times d}$ to compute the first-order proximity:
    $p^{(1)}(v_i^{[\alpha]}, v_j^{[\alpha]}; \zZ, \mathbf{R}) = \sigma\left(\big(\rr_\alpha+\zz_{h,i}^{[\alpha]}\big)^\intercal \big(\zz_{t,j}^{[\alpha]}\big)\right)\:.$
Furthermore, \cite{MELL} regularize the objective function to minimize the variances of the tail and head embedding matrices, to encourage embeddings from replica nodes to be more similar, akin to the consensus embeddings in MTNE. Then, the embeddings obtained with MELL belong to the class of numerous embeddings.

Another extension of LINE to multiplex networks is MVE~\citep{MVE}, which assumes  that $p^{(2)}(v_i^{[\alpha]},v_j^{[\alpha]}; \zZ, (\textbf{c}_i)_{v_i \in V})\propto \exp(\cc_i^\intercal \zz_j^{[\alpha]})$. Then the softmax value is heuristically approximated using the Skip-Gram with negative sampling objective~\citep{SkipGram} in Equation~\eqref{eq:SkipGram}. Finally, the objective function is regularized to force collaboration across the layers, with the term $   \sum_{\alpha\in Y}\sum_{v_i\in V}\lambda_i^\alpha\|\zz_i^{[\alpha]}-\zz_i\|_2^2\:,$ where $\lambda_i^\alpha$ are the weights indicating the influence of each node in each layer, and $ \zz_i$ is the unique cross-layer embedding associated with node $i$, of the form $  \zz_i = \sum_{\alpha\in Y}\lambda_i^\alpha \zz_i^{[\alpha]}\:$. The $(\lambda_i^\alpha)_{v_i \in V,\alpha \in Y}$ are learned via an attention mechanism, using a version of the softmax function and optimized over the cosine similarities of the embedding vectors. \cite{MVE} found that their MVE model computes embeddings efficiently, performing comparably to LINE~\citep{LINE} and node2vec~\citep{node2vec}.

Finally, another approach for multiplex networks which is distinct from LINE-like methods is MUNEM~\citep{MUNEM}. At each epoch,~\cite{MUNEM} sample triplets of nodes: an anchor node, a connected node, and an unconnected node, retrieved by corrupting the source node, and simultaneously optimize layer-specific and cross-layer objective functions over embeddings. The layer-specific objective function ensures that connected nodes are embedded closer to each other than unconnected nodes: $\mathcal{L}^\text{(lay)}(\zZ_\alpha) = \sum_{(v^{[\alpha]}_a,v^{[\alpha]}_p,v^{[\alpha]}_n)\in \mathcal{T}^\alpha}{\max\left\{0,\|\zz^{[\alpha]}_a-\zz^{[\alpha]}_p\|_2^2-\|\zz^{[\alpha]}_a-\zz^{[\alpha]}_n\|_2^2+\delta_{in}\right\}}\:,$ where $\mathcal{T}^\alpha$ is the set of sampled triplets in layer $\alpha$, and $\delta_{in}$ is a hyperparameter controlling the expected similarity between embeddings of connected versus unconnected nodes. The cross-layer objective function promotes collaboration among the different layers by sampling a triplet consisting of an anchor node, its replica node, and a unconnected node on the same layer: $\mathcal{L}^\text{(cross)}(\zZ_1,\dots,\zZ_M) = \sum_{(v^{[\alpha]}_a,v^{[\beta]}_a,v^{[\beta]}_n)\in \mathcal{T}}{\max\left\{0,\|\zz^{[\alpha]}_a-\zz^{[\beta]}_a\|_2^2-\|\zz^{[\alpha]}_a-\zz^{[\beta]}_n\|_2^2+\delta_\text{bet}\right\}}\:,$ where $\mathcal{T}$ is the set of the aforementioned sampled triplets, and $\delta_\text{bet}$ is again a margin hyperparameter. Then, the final objective function in MUNEM is given by
\begin{equation*}
    \mathcal{L}(\zZ_1,\dots,\zZ_M) = \mathcal{L}^\text{(cross)}(\zZ_1,\dots,\zZ_M) + \sum_{\alpha\in Y} \mathcal{L}^\text{(lay)}(\zZ_\alpha)\;.
\end{equation*}
\cite{MUNEM} acknowledge that sampling the entirety of possible connections is intractable, and suggest sampling only a portion of the links for computational efficacy, although they do not elaborate on the sampling scheme or the sampling size that should be used.

Most optimization methods generally operate on the definition of first-order and  second-order proximities on the node pairwise comparison on embeddings. As illustrated in this section, this comparison typically resorts to the sigmoid function when using a probability-based approach, and a normed distance otherwise. 

\subsubsection{Neural Network-Based Methods}\label{neraul-network}

Artificial neural networks have become an exceptionally powerful and versatile framework for learning from data~\citep{DeepLearningBook}. In particular, graph neural networks  are neural networks applied to the graph domain~\citep {GNNReviewZhou}. Similarly to~\cite{MultiplexClusterReview}, we notice that some researchers attempt to extend single-layer neural network approaches to multiplex scenarios. Such approaches can be sorted into three categories~\citep{GNNTutorial}: recurrent graph neural networks, graph convolution neural networks with spatial and spectral approaches, and graph autoencoders, including graph adversarial techniques.

\paragraph{Graph Convolution Neural Networks.} 
Graph convolutional neural networks (GCNs) resort to filtering around a neighborhood of nodes to compute node representations, in a manner similar to classical convolutional neural networks. At each iteration, the representation of a node is updated with the aggregation of the representations of neighboring (connected) nodes. Hence, graph neural networks inherit desirable properties from convolutional neural networks: locality, scalability, and interpretability~\citep{GNNTutorial}. In~\cite{WhereFuseInformation}, the corresponding aggregation step is called a GNN-level information fusion. 

Unlike any of the other models presented in this paper, GNNs require an additional feature matrix $\xX\in\R^{N\times F}$ for nodes. Then the triplet $\mathcal{M}=(Y,\mathcal{G}, \xX)$ defines an \textit{attributed} multiplex network, such that a feature vector $\xx_i\in\R^{F}$ is assigned to every node $v_i\in V$. The node feature matrix is typically used to initialize the recurrent computation of the embeddings, \ie the node feature vectors $\xx_i\in\R^F$ are set as the initial embeddings $\zz^{(0)}_i=\xx_i$ and the resulting final embedding is typically set to the unique embedding of the node $\zz_i=\zz^{(L)}_i$, for a prespecified number of neural network layers $L\geq 0$. In what follows, we will distinguish between the multiplex network layer, which we will continue to denote by Greek letters, and neural network layers (NN-layer), which will be denoted by the index $\ell$. Hence, the embedding of node $v_i$ at NN-layer $\ell$ is given by $\zz^{(\ell)}_i$. The formulation for the NN-layers is often written in terms of the embedding matrix, which is represented by $\zZ^{(\ell)}=[\zz^{(\ell)}_1,\dots, \zz^{(\ell)}_N]^{\top}$ to simplify notation.

So far, all graph convolutional network (GCN)-based approaches for multiplex network embedding rely on two prominent monoplex models: GraphSAGE~\citep{GraphSAGE} and GAT~\citep{GAT}. 

GraphSAGE introduced a two-step process to compute the representation vectors for any node $v_i$ at each NN-layer $\ell$. First, a neighborhood embedding vector $\zz^{(\ell)}_{\mathcal{N}(v_i)}$ is computed via an aggregation method applied to the neighborhood $\mathcal{N}(v_i)$ of node $v_i$. Then $\zz^{(\ell)}_{\mathcal{N}(v_i)}$ and the embedding $\zz^{(\ell-1)}_i$ of node $v_i$ at NN-layer $\ell-1$ are jointly transformed, and run through a nonlinear activation function. Formally, for a node $v_i$ at NN-layer $\ell$, the steps are given by
\begin{align*}
    & \text{(1) } \zz^{(\ell)}_{\mathcal{N}(v_i)} = \text{Agg}_\ell\left(\big\{\zz^{(\ell-1)}_j;\, v_j\in\mathcal{N}(v_i)\big\}\right)\\
    & \text{(2) } \zz^{(\ell)}_{i} =  \text{act}_\ell\left(\wW_\ell\cdot \text{concat}\left(\zz^{(\ell-1)}_i, \zz^{(\ell)}_{\mathcal{N}(v_i)}\right)\right),
\end{align*}
where $\text{Agg}_\ell$ is the differentiable aggregation method for NN-layer $\ell$, $\text{act}_\ell$ is the activation function at NN-layer $\ell$, $\text{concat}$ is the operator for concatenation, and finally, $\wW_\ell\in\R^{d_\ell\times 2d_{\ell-1}}$ is the weight matrix for NN-layer $\ell$ with $d_\ell$ the dimension of the $\ell^\text{th}$ NN-layer. Then, \cite{GraphSAGE} give three examples of aggregation functions: the mean aggregator, the LSTM aggregator, and the pooling aggregator.

~\cite{GAT} take a different approach, they design a graph attention mechanism that learns the importance of each node within its corresponding neighborhood. Their method GAT achieves this by first running a self-attention NN-layer on each node embedding, and then, for each edge $(v_i,v_j)$, taking the softmax of the attention coefficients over all edges involving $v_i$.
\begin{equation*}
    \lambda_{i,j} = \text{Softmax}\left(\text{att}\left(\wW\zz_i,\wW\zz_j\right)\right) = \frac{\exp\left(\text{att}\left(\wW\zz_i,\wW\zz_j\right)\right)}{\sum_{v_k\in\mathcal{N}(v_i)}{\exp\left(\text{att}\left(\wW\zz_i,\wW\zz_k\right)\right)}}\:,
\end{equation*}
where $\wW\in\R^{d\times F}$ is the weight matrix of the attention mechanism, and $\text{att}$ is the attention function. Then, the resulting embedding for node $v_i$ is given by $\zz_i^{(1)}=\text{act}\left(\sum_{v_j\in\mathcal{N}(v_i)}\lambda_{ij}\wW\zz_j\right)$. To stabilize training, the authors suggest computing the self-attention layer multiple times independently to average the attention. They refer to this procedure as multi-head self-attention. 

GraphSage and GAT have been extended to multiplex networks through MultiplexSAGE~\citep{MultiplexSAGE} and MGAT~\citep{MGAT}, respectively. 

In MultiplexSAGE, \cite{MultiplexSAGE} implement collaboration by using an aggregation function to fuse intra- and interlayer neighborhood information into the node representation at NN-layer $\ell$. That is
\begin{equation}\label{MultiplexSAGEaggregation}
\zz^\ell_i = \text{Agg}_\ell\left(\left\{\zz_j^{(\ell-1)}|\,\forall v_j\in\mathcal{N}_H(v_i)\right\}\cup \left\{\zz_j^{(\ell-1)}|\,\forall v_j\in\mathcal{N}_V(v_i)\right\}\cup \{\zz_i^{(\ell-1)}\}\right)\:,
\end{equation}
where $\mathcal{N}_H(v_i)$ and $\mathcal{N}_V(v_i)$ are the intra- and interlayer neighborhoods of $v_i$. Finally, the embedding of node $v_i$ at NN-layer $\ell$ in MultiplexSAGE is given by
\begin{equation}\label{MultiplexSAGEEquation}
    \zz^{(\ell)}_i = \text{act}    \left(\wW^\ell_H\cdot\sum_{v_j\in\mathcal{N}_H(v_i)}\frac{\zz_j^{(\ell-1)}}{|\mathcal{N}_H(v_i)|}+\wW^\ell_V\cdot\sum_{v_j\in\mathcal{N}_V(v_i)}\frac{\zz_j^{(\ell-1)}}{|\mathcal{N}_V(v_i)|}+ \sS^\ell \zz_i^{(\ell-1)}\right),
\end{equation}
where $\wW^\ell_H$, $\wW^\ell_V$, and $\sS^\ell$ are the respective weight matrices for the NN-layer $\ell$. However, MultiplexSAGE is more appropriate for multilayer networks than for multiplex networks, since the last two terms in the sum of Equation~\eqref{MultiplexSAGEEquation} boil down to $(\wW^\ell_V+\sS^\ell)\zz^{(\ell-1)}_i$ for multiplex networks~\citep{MultilayerReview2014}. Moreover, Equation~\eqref{MultiplexSAGEaggregation} can also be simplified to $\zz^{(\ell)}_i = \text{Agg}\left(\left\{\zz_j^{(\ell-1)}|\,\forall v_j\in\mathcal{N}_H(v_i)\right\}\cup\{\zz_i^{(\ell-1)}\}\right)$ in multiplex networks. This considerably differs from GraphSAGE, because the embedding from the previous NN-layer is also included in the aggregation step in MultiplexSAGE. The experiments in~\cite{MultiplexSAGE} show that GraphSAGE generally outperformed its multiplex counterpart for intralayer link prediction, emphasizing the importance of good integration of the inter- and intralayer information. More refined extensions of GraphSAGE to multiplex networks have been developed to overcome this issue. For instance, CGNN~\citep{CGNN} uses a very similar approach to MultiplexSAGE but is restricted to two-layered multiplex networks. In matrix form, their aggregation step is written as 
\[\zZ_\alpha^{\ell}=\text{act}\left(\zZ_\alpha^{(\ell-1)}\wW_1^\ell+\dD_\alpha^{-1/2}\aA_\alpha \dD_\alpha^{-1/2}\zZ_\alpha^{(\ell-1)}\wW^\ell_2+\bB^\ell \zZ^{(\ell-1)}_\beta \wW^\ell_{3}\right)\;,\] 
where the elements of the matrix $\bB^\ell$ are given by $[\bB^\ell]_{ij}=\sigma(\zz_i^{(\ell-1)\intercal}\zz_j^{(\ell-1)})$, with $\mathbf{B}^0$ being the null matrix. Then, the parameters in CGNN are optimized over an intralayer information-related objective function, given by the sum of the layer Skip-Gram objective (Equation~\eqref{eq:SkipGram}) and an interlayer objective function. This objective function pushes embeddings of replica nodes closer together while pulling embeddings of non-replica nodes apart using the cosine similarity.

\begin{figure}
    \centering
    \includegraphics[width=0.9\linewidth]{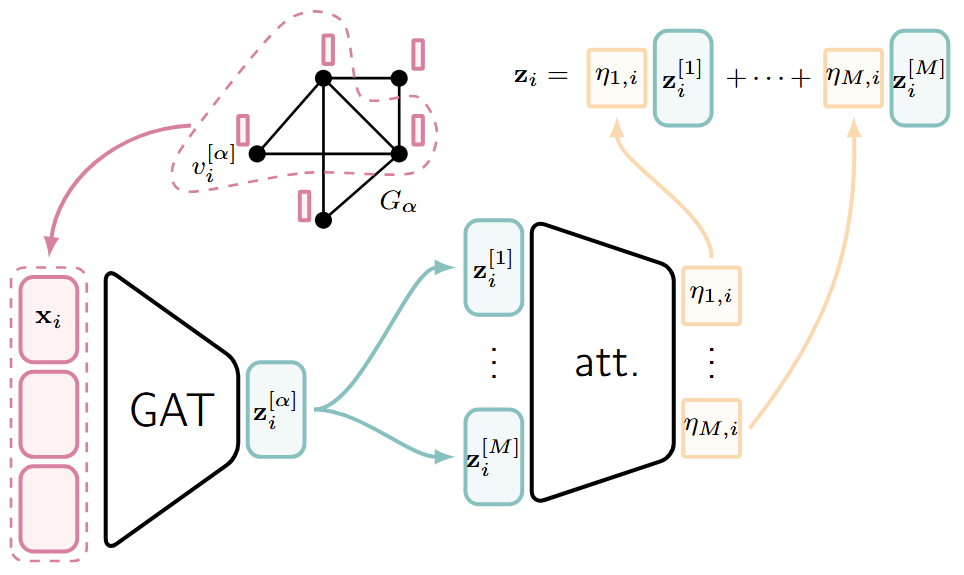}
    \caption{The MGAT algorithm~\citep{MGAT}. First, the 1-step neighbors to node $v_i$ are retrieved. Then, the attribute vectors from all neighbors are merged into a layer-specific representation $h_i^{[\alpha]}$. Using an attention mechanism, the relevance of each layer-specific representation is concentrated in the $\eta_{\alpha,i}$ coefficients, which are the coefficients of the linear combination of neighbor embeddings yielding the final node embedding. 
    }
    \label{fig:mgat}
\end{figure}

On the other hand, MGAT~\citep{MGAT} extends GAT~\citep{GAT} to a multiplex context. MGAT computes a layer-specific multihead self-attention mechanism, which forces the weight matrices $(\wW_\ell)_\ell$ to be similar across NN-layers $k,\ell$ by including the sum over all possible weight differences $\|\wW_k-\wW_\ell\|_2^2$ in the objective function. To construct a unique node embedding from these layer-specific representations, they again use an attention mechanism, where a coefficient $\eta_{\alpha,i}$ associated with layer $\alpha$ and node $v_i$ is computed as
$\eta_{\alpha, i}=\text{Softmax}\left(\ttt_{\alpha}^\intercal (\text{concat}_{\alpha\in Y}\zz^{[\alpha]}_i)\right)$ over the set of replica nodes of $v_i$ in the multiplex graph. In this equation, $\text{concat}_{\alpha\in Y}\zz^{[\alpha]}_i\in\R^{Md}$ is the concatenation of the layer-specific embeddings, and $\ttt_\alpha\in\R^{Md}$ is a trainable parameter for layer $\alpha$. Then, unique node
embeddings are retrieved by taking the linear combination of the node $v_i$'s representations across layers by $(\eta_{\alpha, i})_\alpha$'s. Figure~\ref{fig:mgat} displays the architecture of MGAT.

In a related work, named LIAMNE~\citep{LIAMNE}, the layer attention coefficient is computed as $a_i = \mathbf{w}_1^{\intercal}\tanh\left(\wW_2\cdot\text{concat}_{\alpha\in Y} \zz_i^{[\alpha]}\right)$, where $\mathbf{w}_1\in\R^{Md}$ and $\wW_2\in\R^{N\times Md}$ are trainable parameter vectors and matrices, respectively. The $\zz_i^{[\alpha]}$ in the equation are computed using the GraphSAGE mean aggregator method to node $v_i$ in layer $\alpha$. The final enriched embeddings are retrieved by a learnable affine combination of the final layer-specific embedding $\zz_i^{[\alpha],(L)}$ and the $\text{concat}_{\alpha\in Y} \zz_i^{[\alpha]}$. Moreover, \cite{LIAMNE} also discuss the problem of handling \textit{layer imbalance} on multiplex networks, that is, the discrepancy in the number of edges across layers. This discrepancy may lead to learning bias and performance degradation on the most sparse layers. \cite{LIAMNE} addresses this issue by first constructing a new multiplex network through a sampling scheme that retains the full target layer but undersamples auxiliary layers, on which LIAMNE is applied. 

mGCN~\citep{mGCN2} takes a different approach to implementing the layer attention mechanism than MGAT and LIAMNE. For each node, mGCN maintains at the same time a layer-specific embedding matrix $\zZ^{(\ell)}_\alpha$ for each layer $\alpha$ and a common embedding matrix  $\zZ^{(\ell-1)}$ across layers, and then alternatingly updates them. The layer-specific embedding matrix is obtained from the common embedding matrix with the operation $\zZ^{(\ell)}_\alpha=\text{act}\left(\wW_\alpha \zZ^{(\ell-1)}\right)$; the common embedding matrix is conversely updated with $\zZ^{(\ell)}=\text{act}\left(\wW \text{concat}_{\alpha\in Y}\zZ_\alpha^{(\ell-1)}\right)$. Moreover, $\zZ^{(\ell)}_\alpha$ is updated for each layer $\alpha$ by the convex combination of $\zZ^{(\ell)}_\alpha$ multiplied by the row-normalized adjacency matrix with self-loops $\hat{\aA}_\alpha=\dD_\alpha^{-1}(\aA_\alpha+\iI)$, and the result of the attention mechanism with the attention function $\text{tr}\left(\wW_\alpha^\intercal \mathbf{M} \wW_\beta\right)$. Note that the attention mechanism is applied to the weight matrices $\wW_\alpha, \wW_\beta$ instead of the intermediate embeddings $\zz^\ell_\alpha$, as in previous models. 

Among the methods described in this section, we could have made a distinction between spatial and spectral methods. However, we believe our presentation enables us to better capture the similarities among the different models. Note that models using the structure $\zZ^{(\ell)} = \text{act}(\hat{\aA}\zZ^{(\ell-1)}\wW)$ are all linear approximations of Chebyshev polynomial filters in the spectral domain, with a constrained number of free parameters~\citep{KipfAndWelling}. As such, they demonstrate an equivalence with the spatial approach described, for instance, in~\cite{GNNTutorial}. The renormalization $\hat{\aA}$ of the adjacency matrix with self-loops is a simplification of the technically correct result $\iI + \dD^{-1/2}\aA\dD^{-1/2}$. This simplification avoids numerical instabilities as a result of exploding or vanishing gradients during the optimization process~\citep{Defferrard}.

\paragraph{Adversarial Networks} Generative adversarial networks (GANs) implement a \textit{minimax} game in which a generator, typically referred to with the letter $\mathbb{G}$, generates data to trick a discriminator, referred to with the letter $\mathbb{D}$, which determines whether the data is true or not. 

The application of adversarial networks to representation learning on monoplex networks was initiated by GraphGAN~\citep{GraphGAN}. GraphGAN implements a generator that produces node embeddings, and a discriminator for inferring whether an edge is present in the network using the generated embeddings. The probability of an edge $(v_i,v_j)$ is the product of the sigmoid function applied on the dot product of the corresponding embeddings $\zz_i$ and $\zz_j$: $\mathbb{D}(v_i,v_j)=\sigma(\zz_i^\intercal \zz_j)$. However, the definition of the generator is more sophisticated. \cite{GraphGAN} construct a more computationally efficient graph-specific approximation of the softmax operator, which they refer to as Graph-Softmax. Finally,  the goal is to find the respective parameters $\theta_\mathbb{G}$ and $\theta_\mathbb{D}$ of the generator and the discriminator that are solution to the following optimization problem 
\begin{equation}\label{GraphGANObjective}
\min_{\theta_\mathbb{G}} \ \max_{\theta_\mathbb{D}} \ \mathcal{V}(\mathbb{G},\mathbb{D}) := \sum_{i=1}^{N}{\EE_{v\sim p_\star(v_i)}\left[\log \mathbb{D}(v,v_i;\theta_\mathbb{D})\right] + \EE_{v\sim p_\mathbb{G}(v_i;\theta_\mathbb{G})}\left[\log(1-\mathbb{D}(v,v_i;\theta_\mathbb{D}))\right]}\:, 
\end{equation}
where $p_\star(v)$ and $p_\mathbb{G}(v;\theta_{\mathbb{G}})$ are, respectively, the true and the generated connectivity distribution on edges involving node $v$. The solutions to the two-player minimax game are iteratively updated until they converge to a stable solution. However, this convergence is not guaranteed, even if regularization might help~\citep{GANConvergence}. Now, we consider extensions of GraphGAN to multiplex networks.

Since the Graph-Softmax cannot easily be generalized to multiplex networks, the authors of MEGAN~\citep{MEGAN} resorted to another architecture for the generator $\mathbb{G}$. First, a generator fuses the representations of two nodes together, and computes from that representation the layer-specific probability that the corresponding nodes are connected. The product of these probabilities across the multiplex layers is evaluated to obtain a final probability value indicating whether the two nodes are connected. Then, similarly to the Graph-Softmax approach in GraphGAN, MEGAN reduces the computational complexity by restricting negative samples to a neighborhood of the source node. The neighborhood of a node $v_i$ is defined as the set of replica nodes that are connected to $v_i$: $\mathcal{N}(v_i)=\{v_i\in V\,|\, \exists \alpha\in Y: (v_i^{[\alpha]},v_j^{[\alpha]}) \in E_\alpha\}$. The resulting objective function for MEGAN is similar to Equation~\eqref{GraphGANObjective} in GraphGAN.

Another extension, VANE~\citep{VANE}, solves a different adversarial game. VANE applies two simultaneous games, each with three players. The first game includes the players (1) $F_N$, which is a fully connected layer for one-hot encoded vectors; (2) a long-short-term memory model (LSTM)~\citep{LSTM} $F_S$, which takes in the node embeddings of a sequence of nodes $S$ sampled through a multiplex layer-specific random walk, and returns a subgraph representation $\zz_S$; (3) the 
discriminator $\mathbb{D}_S$ takes in the subgraph embedding and predicts the layer the subgraph lies in. Together, $F_N$ and $F_S$ extract multiplex unique node embedding representations. The game is characterized by the optimization problem $\min_F \ \max_{\mathbb{D}_S} \ \EE_{\zz_S\sim p_\alpha(\zz_S)}\left[\log \mathbb{D}_S(\zz_S)\right] + \EE_{\zz_S\sim \bar{p}_\alpha(\zz_S)}\left[\log(1- \mathbb{D}_S(\zz_S))\right]$, where $p_\alpha(\zz_S)$ is the distribution of the subgraph representation for layer $\alpha$ and $\bar{p}_\alpha(\zz_S)$ is the distribution for the layers other than $\alpha$. The second game includes the players $\mathbb{G}$, a generator that constructs fake node representations. $F_N$, the same player as in the first game, is an embedder that helps the discriminator $\mathbb{D}_N$ to distinguish between valid and generated embeddings. The second game is determined by 
\[\min_G \ \max_{F_N} \ \max_{\mathbb{D}_N} \ \EE_{F_N(V)\sim p(F_N(v))}\left[\log \mathbb{D}_N(F_N(v))\right] + \EE_{\zz\sim p_\zz(\zz)}\left[\log(1- \mathbb{D}_N(\mathbb{G}(\zz))\right]\;,\]
where $p_\zz(\zz)$ is a noise distribution and $p$ is the generated distribution. Contrary to MEGAN, which iterates the game until convergence, VANE iterates the game for a predetermined number of sampled node sequences.

A somewhat different approach to classical adversarial games on monoplex networks like GraphGAN is taken in DGI~\citep{DGI}. First, node embeddings $(\zz_i)_{v_i \in V}$ are generated using a GCN. Second, representations $(\widetilde{\zz}_i)_{v_i \in V}$ on a corrupted version of the network and node attributes are obtained through the same GCN. Third, a graph representation $s$ is computed by taking the sigmoid of the average of all true node representations,  $\mathbf{s}=\sigma\left(\frac1{N}\sum_{v_i\in V}\zz_i\right)$. Finally, the discriminator $\mathbb{D}(\zz;\mathbf{s}) = \sigma(\zz^\intercal \wW\mathbf{s})$ takes in a node embedding $\zz$ and outputs the probability of the embedding being valid. The objective function is the cross-entropy loss between predictions made on the true node representation $\zz_i$ and the corrupted one $\widetilde{\zz}_i$ for each node $v_i \in V$. 

A variant of DGI (called MNI-DGI) for multiplex networks is introduced in~\cite{MNI}, which applies DGI layer-wise and defines an objective function called Infomax, that maximizes mutual information to achieve collaboration between the layers. Note that this model does not fall into the category of aggregation methods, because the Infomax approach is not applied retrospectively. Another extension of DGI to multiplex networks is DMGI~\citep{DMGI}. DMGI applies the DGI model separately to each layer of the multiplex and fosters interlayer collaboration through two techniques. First, DMGI defines a single discriminator common to all layers, and second, the authors introduce a consensus matrix-type technique, similar to that introduced in Section \ref{optimization} for the MTNE model~\citep{MTNE}. The loss function (to minimize) is defined as 
\[\mathcal{L}(\kK, \zZ_1,\dots, \zZ_M, \widetilde{\zZ}_1, \dots, \widetilde{\zZ}_M)=\left[\kK-\frac1{M}\sum_{\alpha\in Y}\zZ_\alpha\right]^2-\left[\kK-\frac1{M}\sum_{\alpha\in Y}\widetilde{\zZ}_\alpha\right]^2\;,\] where $\kK\in\R^{N\times d}$ is the consensus matrix across layers, $(\zZ_\alpha)_{\alpha\in Y}$ are the true and $(\widetilde{\zZ}_\alpha)_{\alpha\in Y}$ are the corrupted layer-specific embedding matrices. As a result, the objective function minimizes the disagreement between the consensus matrix and the true embedding matrices, while maximizing the difference with the corrupted node embeddings. \cite{DMGI} also present a version of the DMGI model that utilizes an attention mechanism, as opposed to the consensus approach, where the attention mechanism for node $v_i$ is defined by the softmax function $a_i=\text{softmax}(\ttt_\alpha^{\intercal}\zz_i^{[\alpha]})$. Again, $\ttt_\alpha$ is a trainable parameter for layer $\alpha$. \cite{DMGI} deduce from their experiments that the attention mechanism generally outperforms the consensus approach. Another work building upon DGI, HMNE~\citep{HMNE} exists but is out of scope for our review, as it was never applied to a link prediction task.

As a conclusion, first, most of the models in this section merely extend approaches on monoplex networks to multiplex networks. Second, most of those models use unique embeddings. The only exceptions to this rule are LIAMNE~\citep{LIAMNE}, CGNN~\citep{CGNN} and MNI-DGI~\citep{MNI}, which generate enriched embeddings. Third, several graph convolutional neural networks have been proposed for clustering and classification tasks, including MAGCN~\citep{MAGCN}, MGCN~\citep{MGCN}, Multi-GCN~\citep{Multi-GCN}, and MR-GCN~\citep{MR-GCN}. We claim that, with only minor adjustments, these networks could also be used for link prediction. Fourth, we were unable to find autoencoder models specifically designed for multiplex or attributed multiplex networks, which paves the way for novel interesting research. Finally, many neural network models are not designed for \textit{weighted} multiplex networks, and therefore, their performance in these scenarios cannot be assessed.

\section{Link Prediction Evaluation}\label{metrics}

\begin{figure}[ht!]
    \centering
    \includegraphics[width=\linewidth]{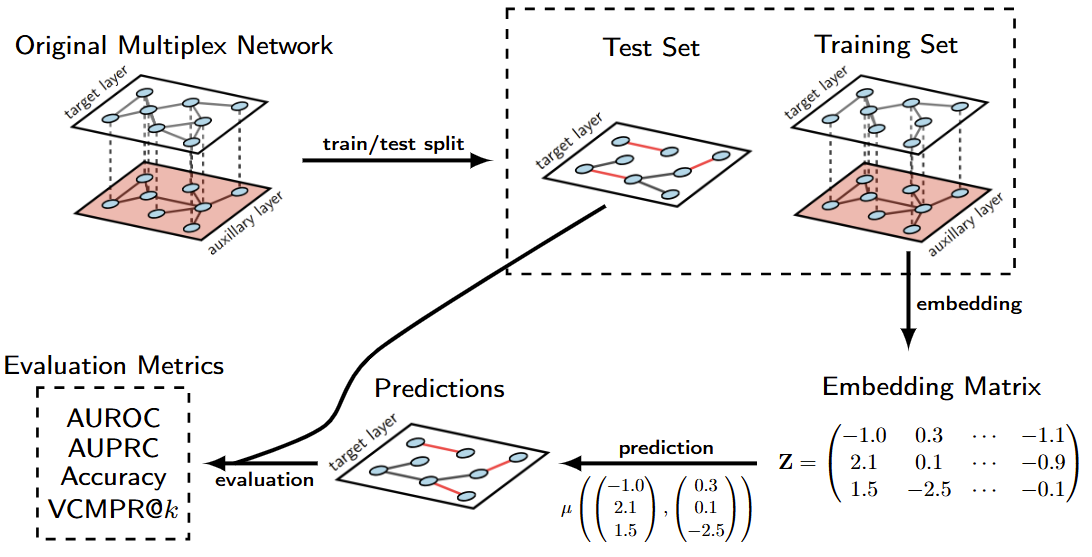}
    \caption{This figure visualizes the evaluation procedure for layer-specific testing procedures. The grey lines in the target layer represent true edges (resp. predicted true edges), whereas the red lines represent the false edges (resp. predicted false edges). The predictor $\mu$ takes in two vector representations, i.e. columns from the embedding matrix $\mathbf{Z}$.}
    \label{fig:evaluation}
\end{figure}

As previously mentioned, link prediction aims to infer edges or estimate their weights in a (multiplex) network. Various factors can influence the performance of link prediction models, and many of these are designed for specific use cases. Consequently, a wide range of evaluation methods has emerged, oe of which is visualized in Figure \ref{fig:evaluation}, making it challenging to compare results across different studies. In this section, we will outline the possible choices for embedding dimensions, prediction methods, evaluation metrics, and testing procedures. We leave out a discussion on the datasets, as the review by~\cite{MultiplexClusterReview} on embedding models for clustering on multiplex networks already describes the most prominent choices for multiplex networks. 

\subsection{Embedding Dimension}

Multiplex networks may exhibit various sizes, ranging from 29 nodes and three layers, as in the Vickers dataset~\citep{vickers}, to over 400,000 nodes and two layers, as in the Twitter dataset~\citep{twitterHiggs}. Criterion \textbf{(iv)} from Section~\ref{RepTaxonomy} requires an efficient embedding model to map the nodes into a low-dimensional latent space, where the dimension $d$ should be chosen to be much smaller than the number of nodes $N$. Typically, the dimension $d$ of the embedding space is set prior to training. Most publications set $d$ to 64 or 128 for large networks~\citep{OhmNet, MEGAN,  MNSF, MultiVERSE, CENALP, MNI, LIAMNE}. Some of the authors argue that fixing the dimension beforehand allows for a better comparison across models~\citep{MELL, mGCN2, MANE+}. In contrast, \cite{gu_principled_2021} mention that the embedding dimension is sometimes also learnt as a hyperparameter for single-layer network approaches. 

However, the choice of $d$ does not seem to be the driving factor for the quality of the latent vector representation, as long as it is large enough. Some authors tested the performance of their model for various sizes, and found no significant improvements beyond $d=16$ or $d=40$~\citep{Multi-node2vec,MNI}. In monoplex networks,  lines, \cite{gu_principled_2021} shows that node2vec~\citep{node2vec} and LINE~\citep{LINE} achieve little to no improvement with dimensions greater than $d=10$ or $d=45$. Interestingly, \cite{FFME&MHME} empirically tested their models FFME and MHME over a range of different embedding dimensions, and observed a significant increase in performance up to $d=100$, although the performance improvement was lesser past $d=50$. The performance of LIAMNE~\citep{LIAMNE} even dropped past $d=64$. 

As a result, some care should be taken when choosing the embedding dimension.~\cite{gu_principled_2021} suggest a principled approach to determine the best embedding size for link prediction in monoplex networks. The lack of theoretical underpinning shows that more research is needed on that front. Moreover, it is unknown how well this approach translates to multiplex networks.

\subsection{Prediction Method}\label{predictionMethod}

After the embedding learning phase, $d$-dimensional node embeddings can be fed to a classifier or a regression tool to make predictions about missing edges in the network. Two approaches are mainly used for predicting intralayer edges: either similarity-based or machine learning-based methods. Similarity-based approaches return a scalar value for each pair of nodes based on their embeddings, whereas machine learning-based methods jointly transform the node embeddings corresponding to the edge, and feed the resulting edge embedding to a downstream ML model.
The expression of the two main prediction methods is provided in Table~\ref{tab:PredictionMethod}.
In either case, the perspective on whether the embeddings represent points or vectors in space is of fundamental importance. Every vector in a Euclidean space is characterized by two points therein; as such, each point can be considered as a vector through the origin. However, one might wonder whether the angle, the specific location in space, or a mixture of these is most relevant for inferring edges. 

\begin{table}[htbp!]
    \centering
    \begin{tabular}{l|c}
    \toprule
        Method & Formula \\
    \midrule
    Cosine Similarity & \(
        \cos(v_i,v_j) = \zz_i^\intercal \zz_j/(\|\zz_i\|\|\zz_j\|)\)\\
    Hadamard Product & \multirow{2}{*}{\(\zz_i\odot \zz_j=([\zz_{i}]_1\cdot [\zz_{j}]_1,\dots,[\zz_{i}]_N\cdot [\zz_{j}]_N)^\intercal\)}\\
    + Classifier/Regressor & \\   
    \bottomrule
    \end{tabular}
    \caption{Main prediction methods for edge presence and edge weight inference in multiplex networks.}
    \label{tab:PredictionMethod}
\end{table}

Among similarity-based approaches, the most common one is the cosine similarity~\citep{MNE, MUNEM, VANE, MGAT, MNI}. Cosine similarity takes the embeddings of two nodes, and computes the cosine value of the angle between them. Therefore, it falls under the methods that neglect the magnitudes of these vectors. The cosine similarity has values in $[-1,1]$, where a value of $1$ (resp., -1) is achieved for collinear vectors of the same (resp., opposite) direction. Hence, a computed value closer to $1$ infers an edge between the nodes. While the cosine similarity is the most popular approach for link prediction in monoplex and multiplex networks, there are growing concerns about its effectiveness due to omitting the information related to the magnitudes~\citep{zhou-etal-2022-problems, CosineSimAboutSim?, you2025semanticsanglecosinesimilarity}. Because of this, the similarities calculated by cosine similarity might be meaningless~\citep{CosineSimAboutSim?}.

Machine learning-based edge prediction methods appear less popular, as they are less frequently cited in papers. Nevertheless, some authors use a logistic regression or a random forest classifier on a transformation of the pair of embeddings to predict corresponding edges~\citep{mGCN2, RMNE, CGNN}. Furthermore, these models usually take the Hadamard product of the two nodes as their input, as first introduced in~\cite{node2vec}, that is, the element-wise multiplication of the two node embeddings associated with the evaluated edge. How the machine learning prediction methods compare to the more popular cosine similarity remains to be assessed. 

Lastly, for the sake of comparability between models, we suggest that the approach should be chosen to work on all types of networks, be it (un)directed and (un)weighted, for the sake of having a reliable predictor on all types of layers.

\subsection{Evaluation Metrics}

\begin{table}[htbp!]
    \centering
    \begin{tabular}{l|c}
    \toprule
        Metric & Formula \\
    \midrule
    Precision & \(\text{TP}/(\text{TP}+\text{FP})\)\\
       % AUPRC & \todoClemence{missing?}\\
       % AUROC & \todoClemence{missing?}\\
        Accuracy & \((\text{TP}+\text{TN})/(\text{TP}+\text{TN}+\text{FP}+\text{FN})\)\\
        F1-Score & \((2\cdot\text{TP})/(2\cdot\text{TP}+\text{FP}+\text{FN})\)\\
        VCMPR@k & \(t_{v_i}(k)/\min\{\deg(v_i), k\}\)\\
        NRMSE & \(\sqrt{\frac1{|D|}\sum_{y\in D}{|\hat{y}-y|^2}}/\max|\hat{y}-y|\)\\
    \bottomrule
    \end{tabular}
    \caption{Metrics for evaluating performance in link prediction. TP: number of true positive edges. FP: number of false positive edges. TN: number of true negative edges. FN: number of false negative edges. }
    \label{tab:EvaluationMetrics}
\end{table}

The most commonly used metrics for link prediction on multiplex networks are the area under the receiver-operator curve (AUROC)~\citep{MVE, MELL, mpx2vec, MTNE,  mGCN2, VANE,MGAT, MNI, LIAMNE}, the area under the precision-recall curve (AUPRC)~\citep{MVN2Vec, MNI}, the average precision~\citep{MEGAN}, the accuracy~\citep{PMNE}, and the F1~\citep{PMNE} scores. The definition of some metrics is recalled in Table~\ref{tab:EvaluationMetrics}. Note that all these metrics are classical performance measures for binary classification problems~\citep{metrics}.

In representation learning in monoplex networks, some researchers have started to question those metrics, and whether they reflect the performance of models in sparse networks. \cite{linkpredictionfaireval} and \cite{VCMPR} argue that the class imbalance--that is, the discrepancy in numbers--between present and absent edges in sparse networks yields a bias in the evaluation using the classical binary classifier metrics. \cite{linkpredictionfaireval} also discuss the difficulty of accurately evaluating directed networks. These concerns also apply to multiplex networks. \citep{VCMPR} concluded that the prediction performance should be evaluated at a vertex level. Both \cite{linkpredictionfaireval, VCMPR} have received little attention so far, yet they emphasize an important issue in the field.

Finally, link prediction on weighted networks has been utterly overlooked to date. A simple approach would use the $\ell_2$-norm, and perform evaluation with the Root Mean Squared Error (RMSE), in an analogy to classical regression problems.

\subsection{Testing Procedure}

Finally, the last crucial step to evaluate an embedding learning model is to construct a testing pipeline. An important part of the evaluation procedure is to construct appropriate training and testing sets from multiplex networks. We can generally distinguish between \textit{general}~\citep{MELL, mpx2vec} and \textit{layer-specific}~\citep{MUNEM, mGCN2, MGAT, MNI} procedures. 

In general procedures, the multiplex network is assumed to be the ground truth. Then, a set of false edges is generated via a random sampling scheme~\citep{MELL, mpx2vec}. In the sampling scheme, false edges $e'$ are sampled uniformly from the set of non-existing edges, \ie  $e' \sim\mathcal{U}\left(\bigcup_{\alpha\in Y}(V_\alpha\times V_\alpha)\setminus (L_\alpha\cup E_\alpha)\right)$, where $\mathcal{U}(\cdot)$ represents the uniform distribution and $L_\alpha=\{(v_i^{[\alpha]},v_i^{[\alpha]})\,|\, i=1,\dots, N\}$ the set of self-loops. Then, the set of all true and false intralayer edges of the network is divided into training, validation, and testing sets. A reduced multiplex network is constructed from the training set of true edges, which is subsequently used to train the embedding learning model. The latent vector pairs for the respective true and false edges of the test set are used to evaluate the prediction method. There are a few adaptations to this procedure.~\cite{VANE} go a step further, and split the data on edges that exist across all layers and replica nodes. In contrast, \cite{mGCN2} adapt the general approach by removing edges between node pairs from the training set at every layer. 

Contrary to the general approach, the layer-specific procedure splits the multiplex network into a \textit{target layer}, whose edges are split into training, validation, and test sets, and \textit{complementary layers} (also known as \textit{auxiliary layers}~\citep{LIAMNE}), which remain unchanged. The complementary layers are used to provide additional information that enhances predictions on the target layer. The false edge sampling scheme for the training and testing sets follows the same procedure as above, although it is limited to the target layer, as depicted in Figure \ref{fig:evaluation}. This procedure offers a significant benefit over the previous one by iteratively choosing each layer in a multiplex network as the target layer, thereby measuring layer-dependent performance. 

Those same procedures have been used to evaluate models on directed networks~\citep{MELL, MGAT}. However, we would like to point out that the likelihood of sampling every edge's reciprocal is small, as most real-world networks exhibit sparsity~\citep{Boccaletti2014TheSA, MultilayerReview2014, BianconiMulNetBook}. We can formally show this by assuming $E\subset V\times V\setminus L$ with $E\neq\emptyset$ and $\#E\ll \#[(V\times V)\setminus L] = \#V\cdot(\#V-1)$, where $L$ is the set of self-loops. Then we define the set of negative reciprocal edges $E_r:=\{(v_i,v_j)\in V\times V\setminus L\,|\,(v_i,v_j)\notin E\text{ and }(v_j,v_i)\in E\}$, for which we clearly have the property $\#E_r \leq \#E$ and $\#L = \#V$. The probability of sampling a reciprocal false edge is now bounded by the probability of sampling a true edge, which is much smaller than sampling from the remaining false edges. In mathematical terms, for an edge $e\in V\times V\setminus L$
\begin{equation*}
    \mathbb{P}(e\in E_r) = \frac{\#E_r}{\#V\cdot(\#V-1)}< \frac{\#(E_r\cup E)}{\#V\cdot(\#V-1)}\ll 1-\frac{\#(E_r\cup E)}{\#V\cdot(\#V-1)} = \mathbb{P}(e\notin E_r\cup E).
\end{equation*}
Since edges in $[(V\times V)\setminus L]\setminus (E\cup E_r)$ have a weight equal to zero in both directions, these edges behave essentially as undirected edges. This effect may artificially inflate the model's performance during evaluation on directed networks when using symmetric predictions. 

Therefore, we suggest instead another two-part testing procedure for directed multiplex networks. In the first test, the false edges are simply the non-existing reciprocal edges. This testing scheme may allow us to determine whether a model can accurately predict the directionality of those edges. The second test samples false edges uniformly from the set $[(V\times V)\setminus L]\setminus (E_r\cup E)$. This scheme may indicate whether using numerous embeddings performs worse than enriched embeddings when the directionality of these false edges is irrelevant.

\section{Conclusion}\label{conclusion}

Embedding methods map nodes to low-dimensional Euclidean spaces to perform downstream tasks such as link prediction. In this paper, we focused on the specific challenges linked to multiplex networks, and extended a model taxonomy for monoplex networks. Then, we leveraged this taxonomy to classify and compare embedding learning models from the literature. We found that many of the models are actually extensions of methods applied to monoplex networks, with additional constraints to achieve collaboration across layers. Furthermore, we introduced a new taxonomy specific to multiplex embedding: the representation taxonomy. This taxonomy enables a more structured comparison of the models. Beyond the taxonomies, we provide an overview of the current approaches to evaluating multiplex network embedding models. Finally, we proposed a new testing procedure for directed multiplex networks, to counteract the shortcomings of the currently used testing schemes.

The field of embedding learning on multiplex networks is still relatively new and remains an actively studied area. Therefore, many avenues of research remain unexplored, ome of which we highlighted in our review, making it a promising field of research with clear real-life applications. 

\backmatter

\section*{Declarations}

\textbf{Conflict of Interest}: The authors have no relevant financial or non-financial interests to disclose.

\noindent\textbf{Funding}: No funding was received to assist with the preparation of this manuscript.

\noindent\textbf{Author Contributions Statement}: O.T. has designed the review and wrote the manuscript's text, tables and figures. O.W. has given substantial proof-reading to the manuscript. C.R. has supervised the review and co-wrote the text. All authors reviewed the manuscript.


\bibliography{references}% common bib file

\end{document}